\PassOptionsToPackage{shortlabels}{enumitem}
\documentclass[11pt, a4paper]{include/gdm_format}

\usepackage{amsmath,amsfonts,bm}

\def\eqref#1{equation~\ref{#1}}

\def\1{\bm{1}}

\DeclareMathAlphabet{\mathsfit}{\encodingdefault}{\sfdefault}{m}{sl}
\SetMathAlphabet{\mathsfit}{bold}{\encodingdefault}{\sfdefault}{bx}{n}

\usepackage[authoryear, sort&compress, round]{natbib}
\usepackage{xcolor}
\usepackage{subcaption}
\usepackage{mwe}
\usepackage{graphicx}
\usepackage{xr}
\usepackage{marvosym}
\usepackage{makecell}
\usepackage[table]{xcolor}
\usepackage{float}

\usepackage{fvextra}

\definecolor{bgcolor}{rgb}{0.95,0.95,0.95}

\definecolor{lblA}{RGB}{33,113,181}
\definecolor{lblB}{RGB}{230,85,13}

\usepackage{tablefootnote}
\usepackage{pdfpages}
\usepackage{etoc}

\usepackage{hyperref}
\usepackage{url}
\usepackage{xurl}
\usepackage[parfill]{parskip}

\usepackage{amsmath}
\usepackage{amsthm}
\usepackage{xspace}
\usepackage{multirow}
\usepackage{paracol}
\usepackage{booktabs}
\usepackage{colortbl}
\usepackage{cleveref}
\usepackage{subcaption}
\usepackage{wrapfig}
\usepackage{soul}
\usepackage{multicol}
\usepackage{listings}
\usepackage{changes}
\usepackage{enumitem}
\usepackage{thmtools}
\usepackage{thm-restate}
\usepackage{graphicx}
\usepackage{adjustbox}
\usepackage{tabularx}
\usepackage{url}
\usepackage{xurl}
\usepackage{hyperref}
\usepackage[parfill]{parskip}

\usepackage{algorithm}
\usepackage{algorithmic}
\usepackage{tcolorbox}

\definecolor{borgogna}{RGB}{159, 29, 53}

\title{Learning from Partial Chain-of-Thought via Truncated-Reasoning Self-Distillation}

\correspondingauthor{Gianluigi Silvestri (\texttt{gianlu.silvestri@gmail.com}), Edoardo Cetin (\texttt{edo@sakana.ai})\\\textsuperscript{*}Work done during an internship at Sakana AI.}

\author[2]{Gianluigi Silvestri\textsuperscript{*}}
\author[1]{Edoardo Cetin}
% ...

\affil[1]{Sakana AI}
\affil[2]{Radboud University}
% ...

\begin{document}
\begin{abstract}
\vspace{-1em}

  Reasoning-oriented language models achieve strong performance by generating long chain-of-thought traces at inference time. However, this capability comes with substantial and often excessive computational cost, which can materialize in redundant or inefficient reasoning. We study this setting and introduce Truncated-Reasoning Self-Distillation (TRSD), a lightweight post-training procedure that encourages models to produce correct predictions from partial reasoning traces. In TRSD, a frozen teacher model first generates a full reasoning trace and evaluates the corresponding answer distribution conditioned on the prompt and the complete reasoning to construct a synthetic training target. A student model with the same architecture is then trained to match the teacher’s answer distribution while being conditioned only on a truncated prefix of its reasoning trace. Across multiple reasoning benchmarks and token budgets, we demonstrate that TRSD improves robustness to truncated inference, with far reduced accuracy tradeoffs when applied to a diverse set of reasoning models. Moreover, although never explicitly regularized for shorter generation during training, we also find that TRSD-trained models inherently output shorter reasoning traces without truncation, significantly reducing inference-time costs even without artificial interventions.
\end{abstract}

\maketitle

\section{Introduction}
\label{sec1:intro}

\begin{wrapfigure}{r}{0.55\textwidth}
\includegraphics[width=\linewidth]{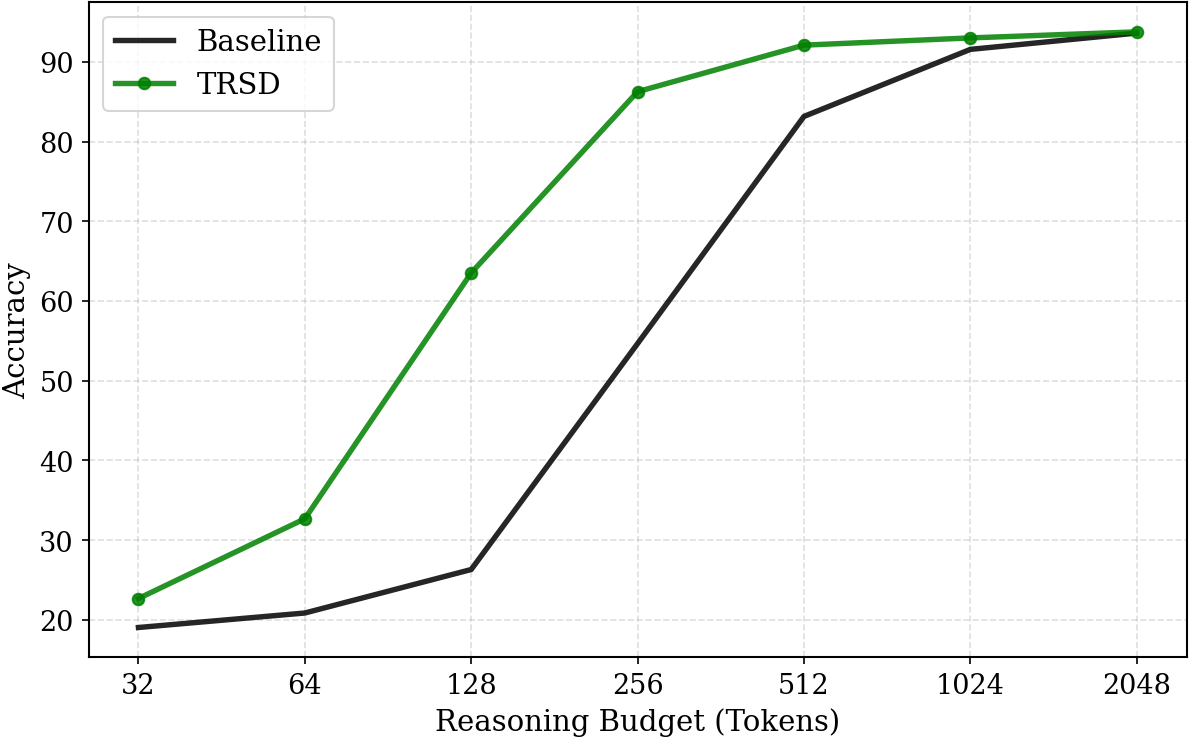}
    \caption{
    Accuracy as a function of the available reasoning budget for a Qwen3-4B model on GSM8K.
    Truncated-Reasoning Self-Distillation (TRSD) substantially improves performance in low-budget regimes, enabling accurate predictions with limited reasoning.
    }
    \label{fig:abstract_trsd_budget}
\end{wrapfigure}
Recent progress in large language models (LLMs) has increasingly focused on solving complex reasoning tasks, moving beyond fluent text generation toward multi-step problem solving in domains such as mathematics and coding \citep{wei2022chain, guo2025deepseek}. A central ingredient behind these gains is the explicit generation of chain-of-thought reasoning, in which models produce long intermediate reasoning traces before emitting a final answer. In modern reasoning models, performance improvements are often achieved by allowing longer such traces at inference time, effectively treating test-time compute as an additional scaling axis alongside model size and training data~\citep{muennighoff2025s1, yang2025qwen3}.

However, this reliance on extended inference comes with significant costs. Reasoning models frequently generate thousands of chain-of-thought tokens per query, resulting in high inference latency and substantial computational overhead. More importantly, this additional computation is often inefficient: models tend to overthink, producing verbose or repetitive reasoning that provides little benefit to the final answer. Prior work shows that such behavior can arise even on simple problems, where extended reasoning is unnecessary and does not reliably improve accuracy~\citep{chen2025do}. In more extreme cases, reasoning traces may degenerate into repetitive or looping patterns, consuming large amounts of inference-time compute without meaningful progress~\citep{pipis2025wait}.

In response to the growing cost of long chain-of-thought reasoning, a range of strategies have been proposed to reduce test-time compute. A first class of approaches acts directly at inference time, guiding the model to produce shorter reasoning traces. This includes prompt-based heuristics, explicit suppression of reflection tokens, and decoding strategies designed to reach a final answer with fewer intermediate reasoning steps~\citep{zhang2023fast, leviathan2025prompt, wang2025wait, xu2502chain}. A second line of work focuses on explicitly compressing or shortening chain-of-thought traces, for instance by skipping, pruning, or selectively retaining reasoning tokens while attempting to preserve final answer accuracy~\citep{yan2025long, xia2025tokenskip, zhang2025lightthinker, zhang2025tokensqueeze, ma2025cot, yuan2025not}. Finally, some methods incorporate length control directly into training, encouraging models to produce shorter reasoning traces by design rather than artificial inference interventions~\citep{luo2025o1, chen2025distilling}.

\begin{figure}[t]
    \centering
    \includegraphics[width=0.98\textwidth]{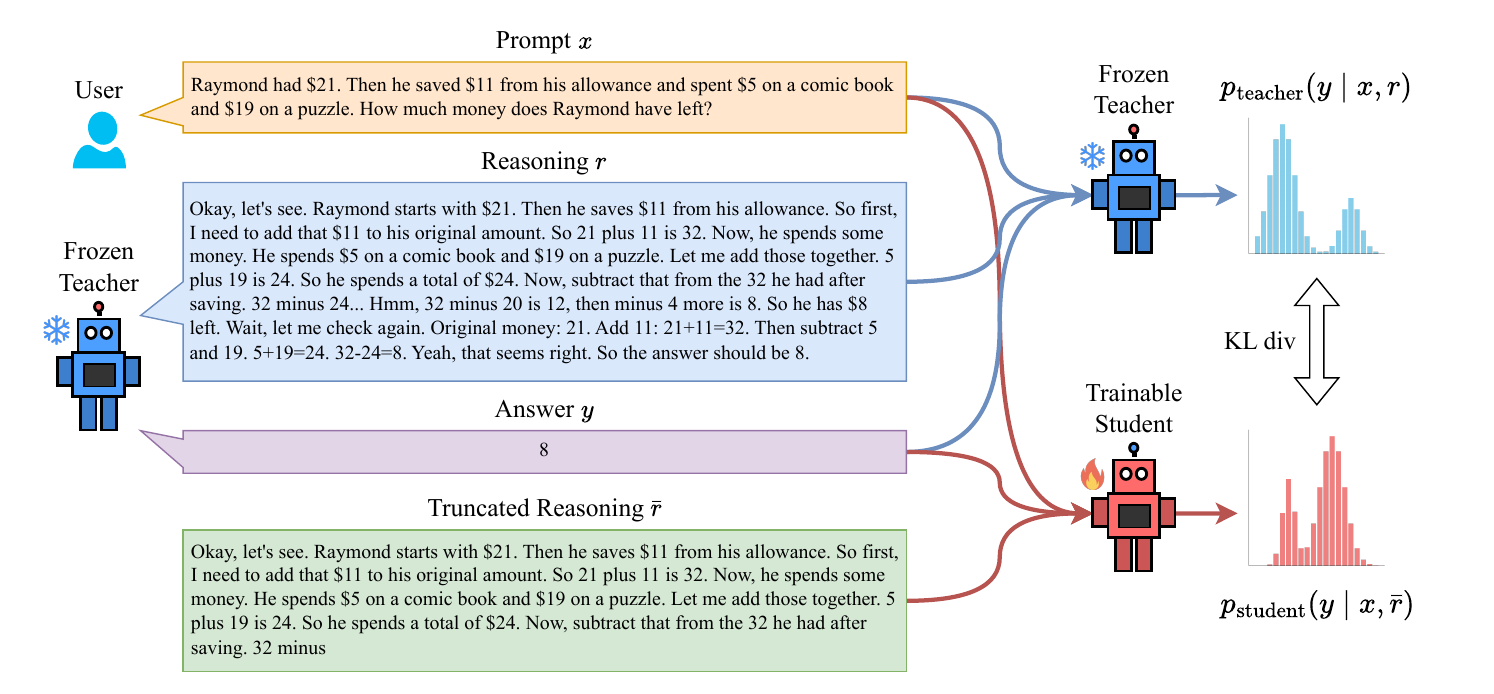}
    \caption{Truncated-Reasoning Self-Distillation (TRSD). 
    Given an input prompt $x$, a frozen teacher model first generates a full chain-of-thought reasoning trace $r$ and an answer $y$, and then evaluates the answer-token distribution $p_{\text{teacher}}(y\mid x,r)$ conditioned on the prompt and the complete reasoning trace. 
    The trainable student model, initialized as a copy of the teacher, is conditioned only on a truncated prefix $\bar r$ of the teacher-generated reasoning trace, and evaluates the corresponding answer distribution $p_{\text{student}}(y\mid x,\bar r)$. 
    Training minimizes the KL divergence between the teacher and student answer distributions, encouraging the student to recover the same predictions from partial reasoning and to remain accurate when inference-time reasoning is truncated.}
    \vspace{-3mm}
    \label{fig:trsd_abstract}
\end{figure}

In this work, we take a different perspective on reducing the cost of reasoning. Rather than modifying decoding procedures or explicitly shortening chain-of-thought traces, we focus on training reasoning models to remain effective when inference-time reasoning is truncated, a setting that naturally arises under latency constraints, compute budgets, or user-facing APIs that allow early stopping. We propose \emph{Truncated-Reasoning Self-Distillation} (TRSD), a post-training self-distillation approach in which a model is trained to match its answer predictions from a truncated prefix of its own reasoning trace to those obtained when conditioned on the full reasoning (see Figure~\ref{fig:trsd_abstract}). By encouraging robustness to truncated reasoning rather than enforcing shorter traces, TRSD directly targets wasteful inference without constraining how models reason, shifting the optimization objective away from producing shorter reasoning and toward producing answers that rely less on extended or redundant computation.

We evaluate TRSD across multiple reasoning-oriented language model architectures and datasets. Across most settings, TRSD-trained models outperform their corresponding baselines under truncated inference (Figure \ref{fig:abstract_trsd_budget}), and in several cases these gains persist even when larger reasoning budgets are available. When reasoning is unconstrained, TRSD-trained models also tend to generate shorter reasoning traces while maintaining or improving accuracy. Importantly, this reduction in reasoning length is not enforced during training, but emerges naturally as a consequence of optimizing robustness to truncated reasoning. Overall, TRSD is a simple and lightweight post-training procedure that improves accuracy under low reasoning budgets and reduces unnecessary inference-time computation in existing reasoning models.
\section{Background}
\label{sec2:background}
Reasoning models are language models trained to explicitly generate chain-of-thought reasoning traces during inference, rather than relying solely on implicit internal computation. This paradigm, exemplified by models such as DeepSeek-R1~\citep{guo2025deepseek}, induces language models to decompose complex problems into intermediate steps by generating long thinking traces before producing a final answer. Training reasoning models typically proceeds in two stages. First, the model undergoes a supervised fine-tuning stage, leveraging masked prompts paired with explicit reasoning traces and correct answers to establish a structured reasoning format in its generation patterns~\citep{wei2022chain, guo2025deepseek}. Second, the model is directly optimized with reinforcement learning to maximize the correctness of its generated responses on downstream tasks using verifiable rewards~\citep{guo2025deepseek}. In this second stage, the thinking trace generated by the language model inherently grows, increasingly trading test-time computation for improved accuracy. To leverage the properties of reasoning while mitigating its inference cost, large reasoning models are often distilled into smaller ``students''. In standard distillation pipelines, a ``teacher'' generates synthetic data consisting of prompts, reasoning traces, and answers, which are used to supervise a smaller model via fine-tuning or distribution matching~\citep{guo2025deepseek, guha2025openthoughts}. While distillation enables compact models to acquire strong reasoning performance~\citep{ye2025limo, muennighoff2025s1}, it can also encourage students to reproduce long and unnecessary reasoning traces, partially inheriting the teacher's inefficiencies~\citep{chen2025do, pipis2025wait, yuan2025not}.

\section{Method}
\label{sec3:method}

\subsection{Truncated-Reasoning Self-Distillation}

We propose \emph{Truncated-Reasoning Self-Distillation} (TRSD), a lightweight post-training procedure applied as an additional fine-tuning step on top of already-trained reasoning models to encourage accurate predictions even with cheaper truncated reasoning traces. 

The core idea is to train a model to match its own predictions under full reasoning while being conditioned on truncated reasoning during training. Let $x$ denote an input prompt, $r$ a generated chain-of-thought reasoning trace, and $y$ the corresponding answer. As in traditional distillation, a teacher model is prompted with $x$ to generate a complete reasoning trace $r$ followed by an answer $y$. However, rather than conditioning on the full reasoning trace, the student model is then trained to match the teacher’s answer distribution while being conditioned only on a truncated prefix of $r$, denoted by $\bar{r}$. The teacher and student share the same architecture and initialization, but while the student is updated, the teacher remains frozen. Thus, all the supervision signal arises from matching the frozen teacher’s predictions under truncated reasoning rather than expensive, larger models. Our design choice of freezing the teacher serves to provide a stable target distribution and allows the student to learn how to recover accurate predictions from its own truncated reasoning, rather than altering the underlying reasoning process itself.

\subsection{Reasoning Truncation}

In TRSD, reasoning traces are truncated at the token level. For each training example, we sample a truncation ratio $\alpha \sim \mathcal{U}(0,1)$,
corresponding to retaining a prefix containing $\lfloor \alpha \cdot |r| \rfloor$ tokens of the original reasoning trace, where $|r|$ denotes the number of tokens in the reasoning trace. Truncation is applied independently for each example in the batch. This simple uniform sampling strategy exposes the student to a broad range of truncated-reasoning regimes, from near-complete traces to very short prefixes. It also reflects the unpredictability of inference-time truncation in real-world settings, where generation may be stopped at arbitrary points.
\subsection{Training Objective}
Following the canonical DeepSeek R1 formulation~\citep{guo2025deepseek}, the teacher model is prompted to generate reasoning within \texttt{<think>} \dots \texttt{</think>} tags, followed by the final answer within \texttt{<answer>} \dots \texttt{</answer>} tags. We enforce a maximum reasoning length with answer forcing~\citep{muennighoff2025s1}: if the model does not emit an answer within a given token budget, we pause generation and append the opening answer tag to elicit immediate answer generation. For each training example, while the teacher generates the answer tokens conditioned on the \emph{full} reasoning trace, the student is trained to match the same answer tokens conditioned on its \emph{truncated} version. To optimize the student, our framework minimizes the KL divergence between the teacher and student answer distributions:
\begin{align}
\mathcal{L}
&= \mathrm{KL}\!\left(
p_{\text{teacher}}(y \mid x, r)
\;\|\;
p_{\text{student}}(y \mid x, \bar{r})
\right).
\end{align}
The loss is computed only over answer tokens, without backpropagation through the teacher (Algorithm~\ref{alg:self-trunc}). This encourages the student to match the teacher’s answer under truncated reasoning, while discouraging reliance on repetitive or late-stage refinement tokens that are often missing. As a result of our design, the model is optimized to prioritize earlier, more informative reasoning tokens, preserving its final response distribution with a cheaper, truncated reasoning trace.

\begin{algorithm}[h]
\caption{Truncated-Reasoning Self-Distillation}
\label{alg:self-trunc}
\begin{algorithmic}[1]
\REQUIRE Prompt $x$, teacher model $f_{\theta_t}$, student model $f_{\theta_s}$, max reasoning length $L$
\ENSURE Updated student parameters $\theta_s$
\STATE $(r, y) \leftarrow f_{\theta_t}(x)$ \quad $\triangleright$ teacher generation (frozen)
\IF{$y$ not produced within $L$}
    \STATE Force answer generation
\ENDIF
\STATE Sample $\alpha \sim \mathcal{U}(0,1)$
\STATE $\bar{r} \leftarrow \text{prefix}(r, \lfloor \alpha |r| \rfloor)$
\STATE $p_T \leftarrow f_{\theta_t}(y \mid x, r)$ \quad $\triangleright$ stop gradient
\STATE $p_S \leftarrow f_{\theta_s}(y \mid x, \bar{r})$
\STATE $\theta_s \leftarrow \arg\min_{\theta_s} \mathrm{KL}(p_T \,\|\, p_S)$
\end{algorithmic}
\end{algorithm}
\section{Experimental Results}
\label{sec4:experiments}

We evaluate TRSD across a wide range of reasoning-oriented language models and training datasets. Specifically, we consider Qwen3 models at 0.6B, 1.7B, and 4B parameters \citep{yang2025qwen3}, Phi-4-mini-reasoning (4B) \citep{xu2025phi}, and OpenThinker-3 (1.5B) \citep{guha2025openthoughts}. For each architecture, models are trained using prompts drawn from a single dataset at a time, spanning a range of reasoning difficulty, including Countdown, GSM8K \citep{cobbe2021gsm8k}, Dolci Math \citep{olmo2025olmo3}, and competition MATH \citep{hendrycksmath2021}. Throughout our analysis, we use the average number of reasoning tokens generated by the baseline model as a coarse proxy for task difficulty, with more challenging tasks typically requiring longer reasoning traces (see, e.g., Table~\ref{tab:thinklen_correct_wrong_combined}). Additional details about training and evaluation datasets are provided in Appendix~\ref{appB:datasets}. To keep the main discussion focused, we concentrate on the two largest models considered in our study, Qwen3-4B and Phi-4-mini-reasoning, both of which have approximately 4B parameters and exhibit the strongest overall performance across benchmarks. Results for smaller models and additional architectures follow the same qualitative trends and are reported in Appendix~\ref{appD:extended_results}, together with extended tables and per-dataset performance plots.

In all experiments, the student model is initialized as an exact copy of the teacher and trained using the procedure described in Section~\ref{sec3:method}. Unless stated otherwise, the teacher is allowed to generate up to 8192 tokens for reasoning. If this budget is exhausted before an answer is produced, we apply answer forcing by appending the end-of-thinking token and allowing an additional 200 tokens for answer generation. The procedure used to process teacher-generated answers is described in Appendix~\ref{app:teacher_answer}, and additional training details are provided in Appendix~\ref{app:training_details}.

At evaluation time, TRSD models autoregressively generate their own reasoning traces without any teacher support. We evaluate performance under a range of fixed reasoning budgets by enforcing a maximum number of reasoning tokens at inference time, after which generation is interrupted. If a model does not emit an answer within the allotted reasoning budget, we apply the same answer forcing procedure used during training, appending the end-of-thinking token and allowing an additional 200 tokens for answer generation~\citep{muennighoff2025s1}.

Throughout this section, we distinguish between \emph{in-distribution} and \emph{out-of-distribution} evaluation with respect to the distribution of prompts used during the self-distillation phase. In-distribution results correspond to models evaluated on the same dataset from which self-distillation prompts were drawn, while out-of-distribution results refer to evaluation on reasoning benchmarks not used during fine-tuning.

\subsection{In-Distribution Performance under Truncated Inference}
We begin by evaluating models on validation sets drawn from the same datasets used during self-distillation, in order to characterize robustness to inference-time truncation in a controlled setting.
Table \ref{tab:in_dist_combined} reports results for Qwen3-4B and Phi-4-mini-reasoning, respectively, across a range of reasoning budgets.

Across both models, TRSD consistently improves performance in heavily truncated regimes. When the available reasoning budget is small (e.g., below 512 tokens), distilled models reliably outperform their corresponding baselines across all datasets. In contrast, at the largest reasoning budgets, performance differences become increasingly trivial and task-dependent, with TRSD students successfully preserving the baseline's performance despite the shift in optimization objective. Crucially, these improvements should not be interpreted as standard task-specific fine-tuning effects. During TRSD, the model is not trained on ground-truth labels from the dataset, nor is it optimized to improve full-budget accuracy on that benchmark. Instead, the student is trained only to reproduce the frozen teacher’s answer distribution under truncated reasoning. The observed gains, therefore, suggest a genuine increase in robustness to limited reasoning budgets, rather than adaptation to the data distribution itself.


\begin{table*}[t]
\centering
\caption{
Baseline vs TRSD under truncated inference for in-distribution prompts.
Entries report Baseline/TRSD accuracy (higher is better), bold indicates the better entry.
When Baseline and TRSD accuracies differ by at most 1 percentage point, both are bold.
}
\label{tab:in_dist_combined}

\resizebox{\columnwidth}{!}{
\begin{tabular}{lcccccccc}
\toprule
\multirow{2}{*}{Budget / Data} &
\multicolumn{2}{c}{Countdown} &
\multicolumn{2}{c}{Dolci} &
\multicolumn{2}{c}{GSM8K} &
\multicolumn{2}{c}{Math500} \\
\cmidrule(lr{4pt}){2-3}
\cmidrule(lr{4pt}){4-5}
\cmidrule(lr{4pt}){6-7}
\cmidrule(l{6pt}){8-9}
& Baseline & TRSD &
Baseline & TRSD &
Baseline & TRSD &
Baseline & TRSD \\
\midrule

\multicolumn{9}{c}{\emph{Teacher Model: Qwen3-4B}} \\
\midrule
32   & 11.72 & \textbf{27.24} \textcolor{LimeGreen}{\scriptsize(\hphantom{-}+15.5)}
     & 8.85  & \textbf{9.90}  \textcolor{LimeGreen}{\scriptsize(\hphantom{-}+1.1)}
     & 19.03 & \textbf{22.67} \textcolor{LimeGreen}{\scriptsize(\hphantom{-}+3.6)}
     & 19.00 & \textbf{23.60} \textcolor{LimeGreen}{\scriptsize(\hphantom{-}+4.6)} \\
64   & 13.46 & \textbf{28.06} \textcolor{LimeGreen}{\scriptsize(\hphantom{-}+14.6)}
     & 8.97  & \textbf{10.54} \textcolor{LimeGreen}{\scriptsize(\hphantom{-}+1.6)}
     & 20.85 & \textbf{32.68} \textcolor{LimeGreen}{\scriptsize(\hphantom{-}+11.8)}
     & 18.60 & \textbf{25.40} \textcolor{LimeGreen}{\scriptsize(\hphantom{-}+6.8)} \\
128  & 26.04 & \textbf{35.12} \textcolor{LimeGreen}{\scriptsize(\hphantom{-}+9.1)}
     & 9.33  & \textbf{11.82} \textcolor{LimeGreen}{\scriptsize(\hphantom{-}+2.5)}
     & 26.31 & \textbf{63.53} \textcolor{LimeGreen}{\scriptsize(\hphantom{-}+37.2)}
     & 21.20 & \textbf{25.80} \textcolor{LimeGreen}{\scriptsize(\hphantom{-}+4.6)} \\
256  & 43.94 & \textbf{49.72} \textcolor{LimeGreen}{\scriptsize(\hphantom{-}+5.8)}
     & 9.94  & \textbf{13.18} \textcolor{LimeGreen}{\scriptsize(\hphantom{-}+3.2)}
     & 54.66 & \textbf{86.28} \textcolor{LimeGreen}{\scriptsize(\hphantom{-}+31.6)}
     & 26.00 & \textbf{35.20} \textcolor{LimeGreen}{\scriptsize(\hphantom{-}+9.2)} \\
512  & 60.28 & \textbf{63.22} \textcolor{LimeGreen}{\scriptsize(\hphantom{-}+2.9)}
     & 11.46 & \textbf{15.46} \textcolor{LimeGreen}{\scriptsize(\hphantom{-}+4.0)}
     & 83.17 & \textbf{92.12} \textcolor{LimeGreen}{\scriptsize(\hphantom{-}+9.0)}
     & 45.40 & \textbf{58.00} \textcolor{LimeGreen}{\scriptsize(\hphantom{-}+12.6)} \\
1024 & 71.14 & \textbf{72.54} \textcolor{LimeGreen}{\scriptsize(\hphantom{-}+1.4)}
     & 13.22 & \textbf{17.19} \textcolor{LimeGreen}{\scriptsize(\hphantom{-}+4.0)}
     & 91.58 & \textbf{93.03} \textcolor{LimeGreen}{\scriptsize(\hphantom{-}+1.4)}
     & 67.00 & \textbf{76.00} \textcolor{LimeGreen}{\scriptsize(\hphantom{-}+9.0)} \\
2048 & 76.64 & \textbf{78.30} \textcolor{LimeGreen}{\scriptsize(\hphantom{-}+1.7)}
     & 17.19 & \textbf{19.87} \textcolor{LimeGreen}{\scriptsize(\hphantom{-}+2.7)}
     & \textbf{93.63} & \textbf{93.78} \textcolor{LimeGreen}{\scriptsize(\hphantom{-}+0.2)}
     & 80.80 & \textbf{85.60} \textcolor{LimeGreen}{\scriptsize(\hphantom{-}+4.8)} \\
4096 & \textbf{81.32} & \textbf{81.20} \textcolor{red}{\scriptsize(\hphantom{+}-0.1)}
     & \textbf{23.32} & 22.72 \textcolor{red}{\scriptsize(\hphantom{+}-0.6)}
     & \textbf{94.31} & 93.86 \textcolor{red}{\scriptsize(\hphantom{+}-0.5)}
     & 88.00 & \textbf{90.00} \textcolor{LimeGreen}{\scriptsize(\hphantom{-}+2.0)} \\
8192 & \textbf{84.88} & 81.30 \textcolor{red}{\scriptsize(\hphantom{+}-3.6)}
     & \textbf{34.21} & 24.16 \textcolor{red}{\scriptsize(\hphantom{+}-10.1)}
     & \textbf{94.54} & \textbf{93.63} \textcolor{red}{\scriptsize(\hphantom{+}-0.9)}
     & \textbf{91.40} & \textbf{91.20} \textcolor{red}{\scriptsize(\hphantom{+}-0.2)} \\

\midrule
\multicolumn{9}{c}{\emph{Teacher Model: Phi-4-mini-reasoning}} \\
\midrule
32   & 4.56 & \textbf{21.76} \textcolor{LimeGreen}{\scriptsize(\hphantom{-}+17.2)}
     & 7.05 & \textbf{9.17}  \textcolor{LimeGreen}{\scriptsize(\hphantom{-}+2.1)}
     & 17.89 & \textbf{23.28} \textcolor{LimeGreen}{\scriptsize(\hphantom{-}+5.4)}
     & 13.40 & \textbf{16.00} \textcolor{LimeGreen}{\scriptsize(\hphantom{-}+2.6)} \\
64   & 7.08 & \textbf{22.58} \textcolor{LimeGreen}{\scriptsize(\hphantom{-}+15.5)}
     & 6.73 & \textbf{9.13}  \textcolor{LimeGreen}{\scriptsize(\hphantom{-}+2.4)}
     & 18.65 & \textbf{35.25} \textcolor{LimeGreen}{\scriptsize(\hphantom{-}+16.6)}
     & 12.80 & \textbf{17.00} \textcolor{LimeGreen}{\scriptsize(\hphantom{-}+4.2)} \\
128  & 10.42 & \textbf{29.68} \textcolor{LimeGreen}{\scriptsize(\hphantom{-}+19.3)}
     & \textbf{8.33} & \textbf{9.25}  \textcolor{LimeGreen}{\scriptsize(\hphantom{-}+0.9)}
     & 25.63 & \textbf{50.64} \textcolor{LimeGreen}{\scriptsize(\hphantom{-}+25.0)}
     & 18.00 & \textbf{21.60} \textcolor{LimeGreen}{\scriptsize(\hphantom{-}+3.6)} \\
256  & 32.60 & \textbf{41.64} \textcolor{LimeGreen}{\scriptsize(\hphantom{-}+9.0)}
     & 9.58 & \textbf{11.18} \textcolor{LimeGreen}{\scriptsize(\hphantom{-}+1.6)}
     & 44.96 & \textbf{69.29} \textcolor{LimeGreen}{\scriptsize(\hphantom{-}+24.3)}
     & 21.00 & \textbf{32.60} \textcolor{LimeGreen}{\scriptsize(\hphantom{-}+11.6)} \\
512  & 43.66 & \textbf{51.60} \textcolor{LimeGreen}{\scriptsize(\hphantom{-}+7.9)}
     & 9.62 & \textbf{12.10} \textcolor{LimeGreen}{\scriptsize(\hphantom{-}+2.5)}
     & 80.74 & \textbf{85.29} \textcolor{LimeGreen}{\scriptsize(\hphantom{-}+4.6)}
     & 35.40 & \textbf{44.40} \textcolor{LimeGreen}{\scriptsize(\hphantom{-}+9.0)} \\
1024 & 54.34 & \textbf{57.48} \textcolor{LimeGreen}{\scriptsize(\hphantom{-}+3.1)}
     & 10.58 & \textbf{14.58} \textcolor{LimeGreen}{\scriptsize(\hphantom{-}+4.0)}
     & \textbf{87.72} & \textbf{86.88} \textcolor{red}{\scriptsize(\hphantom{+}-0.8)}
     & 43.40 & \textbf{52.00} \textcolor{LimeGreen}{\scriptsize(\hphantom{-}+8.6)} \\
2048 & \textbf{59.94} & \textbf{59.08} \textcolor{red}{\scriptsize(\hphantom{+}-0.9)}
     & 13.50 & \textbf{16.99} \textcolor{LimeGreen}{\scriptsize(\hphantom{-}+3.5)}
     & \textbf{90.37} & 87.04 \textcolor{red}{\scriptsize(\hphantom{+}-3.3)}
     & 46.40 & \textbf{60.00} \textcolor{LimeGreen}{\scriptsize(\hphantom{-}+13.6)} \\
4096 & \textbf{61.96} & 59.56 \textcolor{red}{\scriptsize(\hphantom{+}-2.4)}
     & 16.47 & \textbf{19.07} \textcolor{LimeGreen}{\scriptsize(\hphantom{-}+2.6)}
     & \textbf{90.67} & 87.11 \textcolor{red}{\scriptsize(\hphantom{+}-3.6)}
     & 51.20 & \textbf{66.20} \textcolor{LimeGreen}{\scriptsize(\hphantom{-}+15.0)} \\
8192 & \textbf{64.24} & 60.28 \textcolor{red}{\scriptsize(\hphantom{+}-4.0)}
     & \textbf{24.44} & 20.15 \textcolor{red}{\scriptsize(\hphantom{+}-4.3)}
     & \textbf{90.90} & 87.34 \textcolor{red}{\scriptsize(\hphantom{+}-3.6)}
     & 56.20 & \textbf{69.40} \textcolor{LimeGreen}{\scriptsize(\hphantom{-}+13.2)} \\
\bottomrule
\end{tabular}
}
\vspace{-3mm}
\end{table*}

\subsection{Out-of-Distribution Generalization across Prompt Distributions}
\begin{figure*}[ht]
    \centering
    \subfloat[Countdown]{\includegraphics[width=0.44\linewidth]{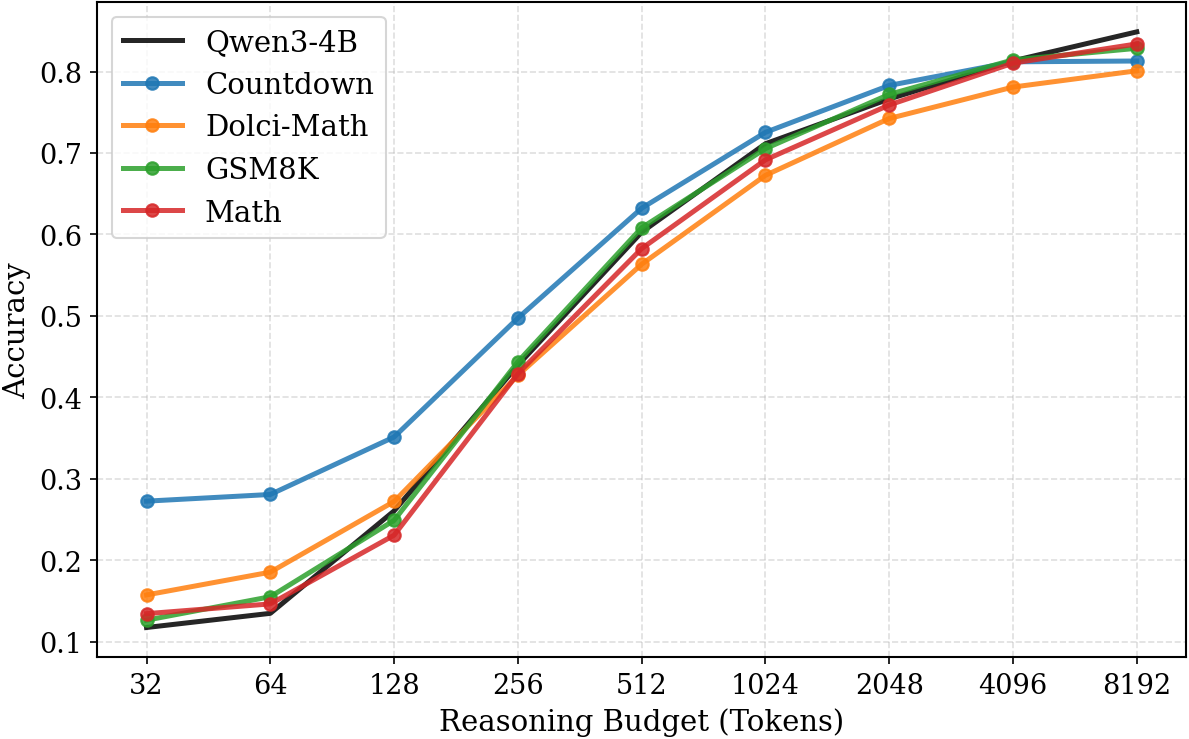}}
    \hfill
    \subfloat[Dolci]{\includegraphics[width=0.44\linewidth]{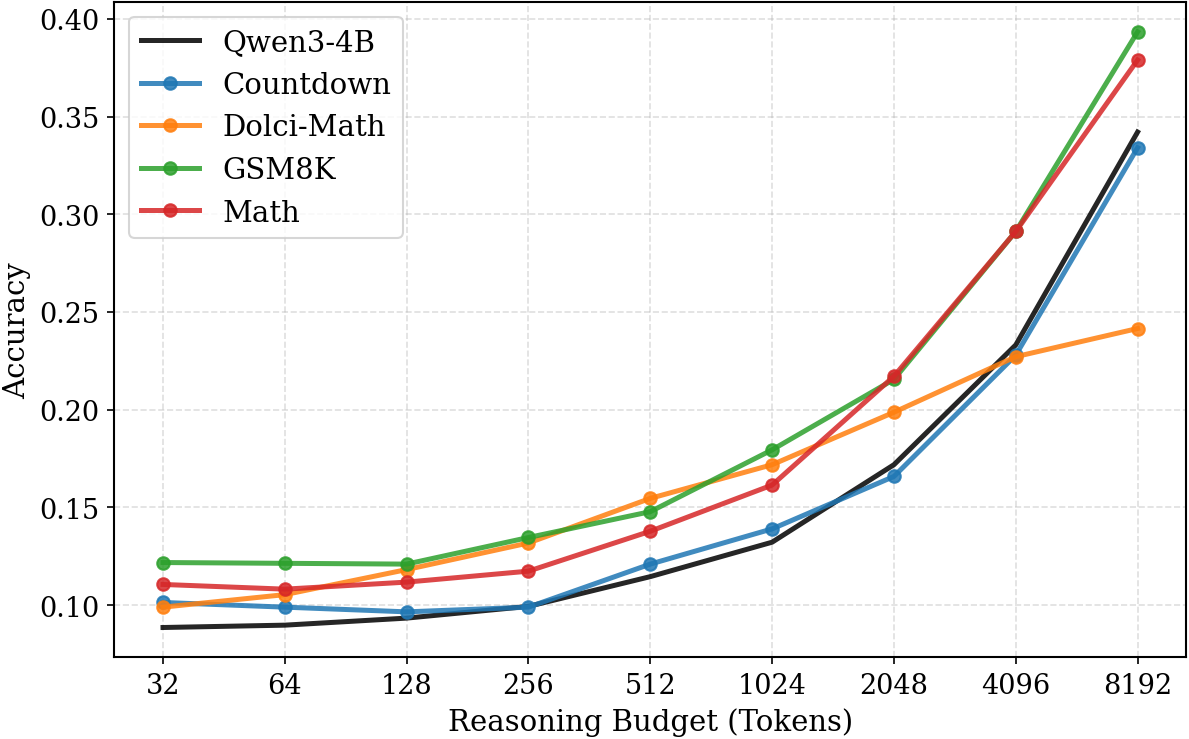}} \\
    \subfloat[GSM8K]{\includegraphics[width=0.44\linewidth]{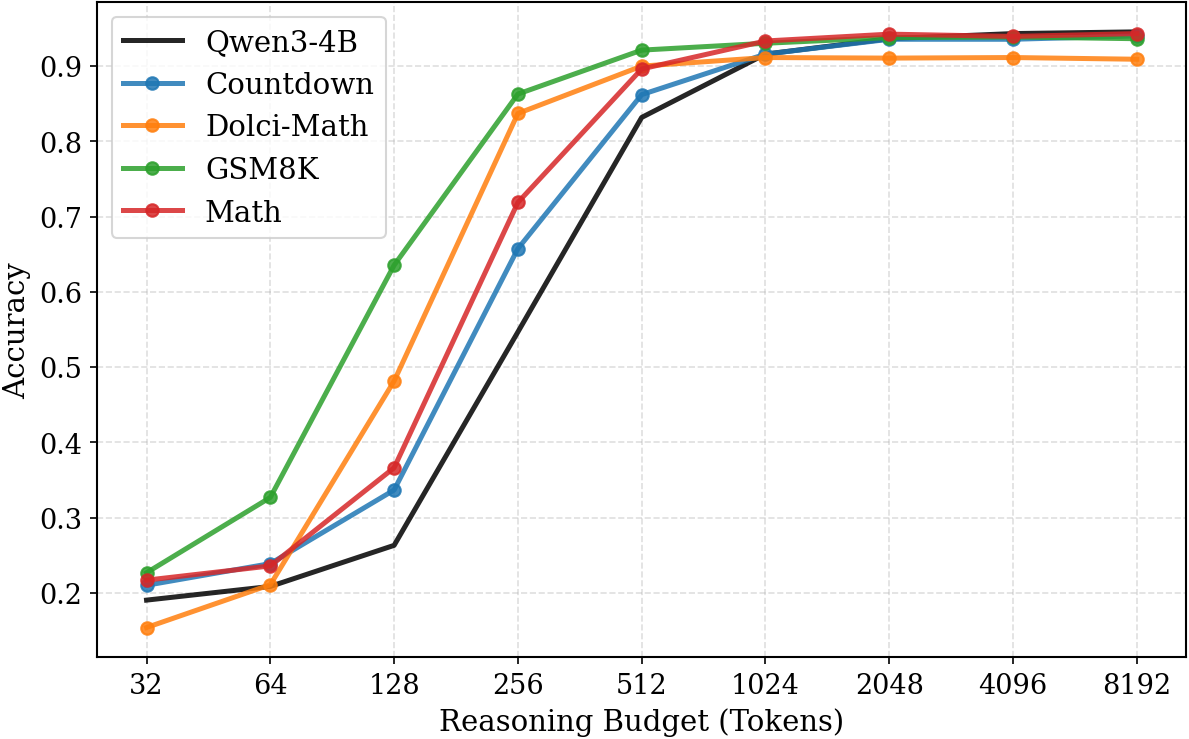}}
    \hfill
    \subfloat[Math500]{\includegraphics[width=0.44\linewidth]{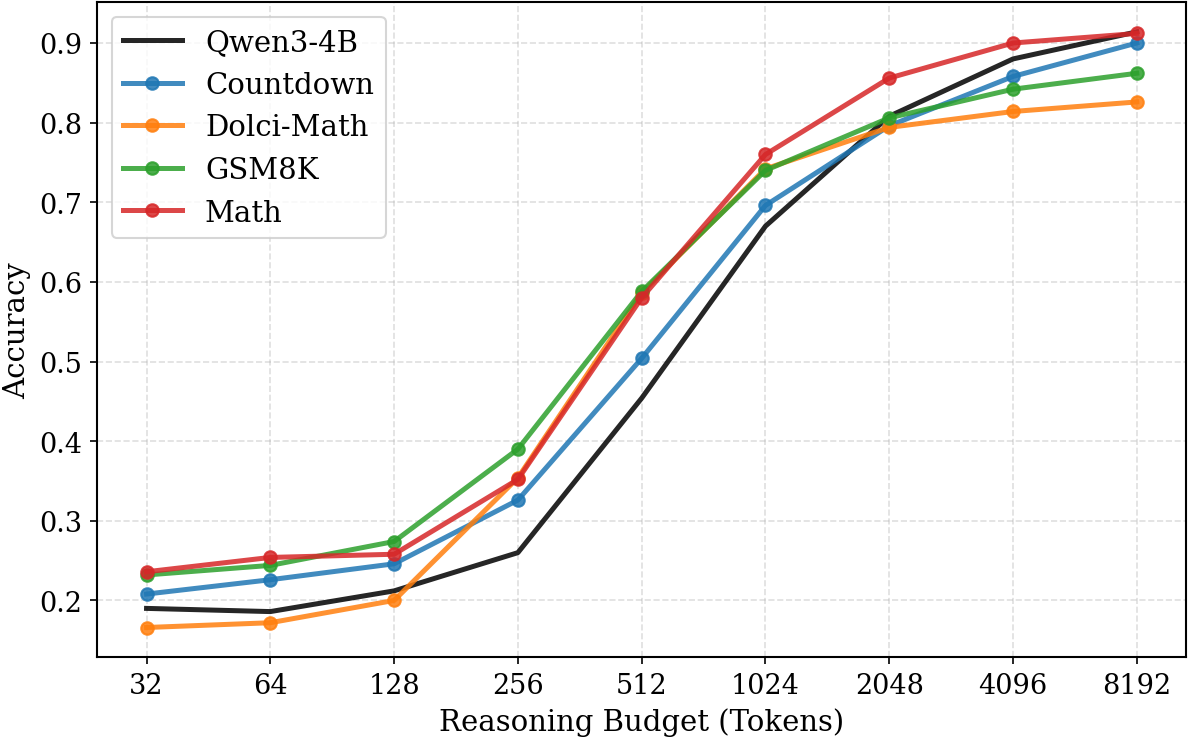}}
    \caption{Per-dataset accuracy as a function of the reasoning budget for Qwen3-4B. The evaluation dataset is specified below the respective plot.}
\label{fig:perdata_qwen3_4b}
\vspace{-3mm}
\end{figure*}

\begin{figure*}[h]
    \centering
    \subfloat[Countdown]{\includegraphics[width=0.44\linewidth]{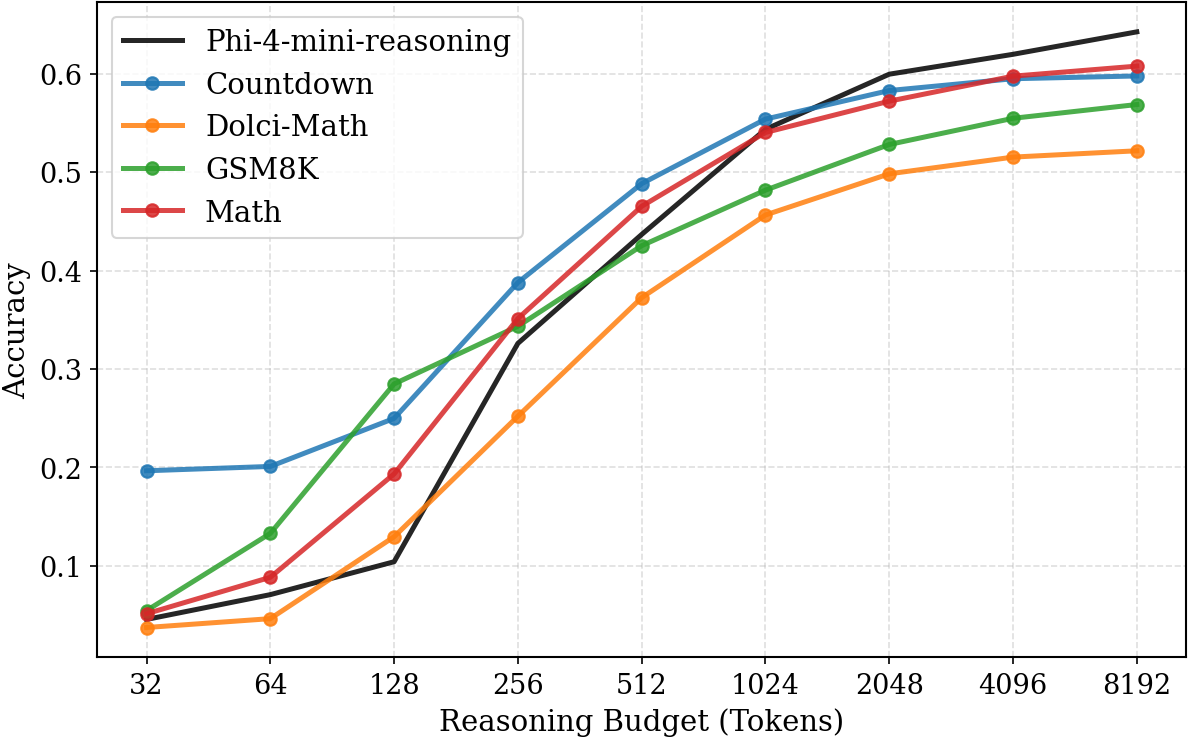}}
    \hfill
    \subfloat[Dolci]{\includegraphics[width=0.44\linewidth]{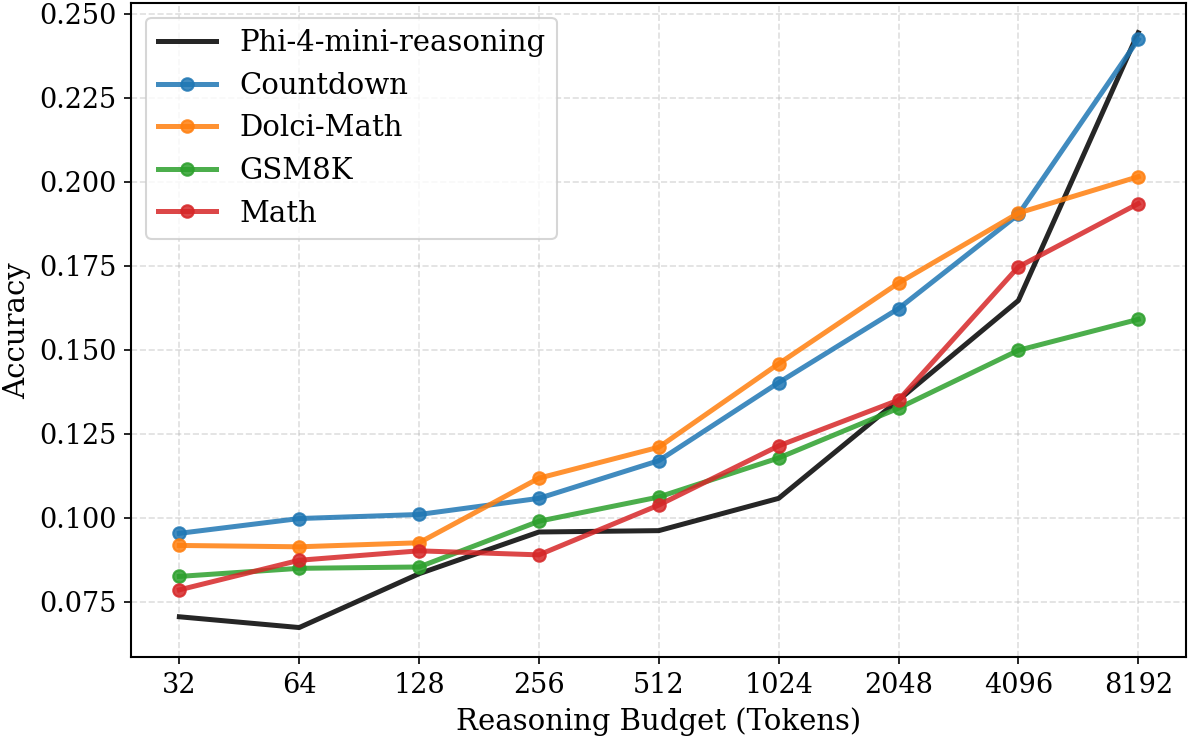}} \\
    \subfloat[GSM8K]{\includegraphics[width=0.44\linewidth]{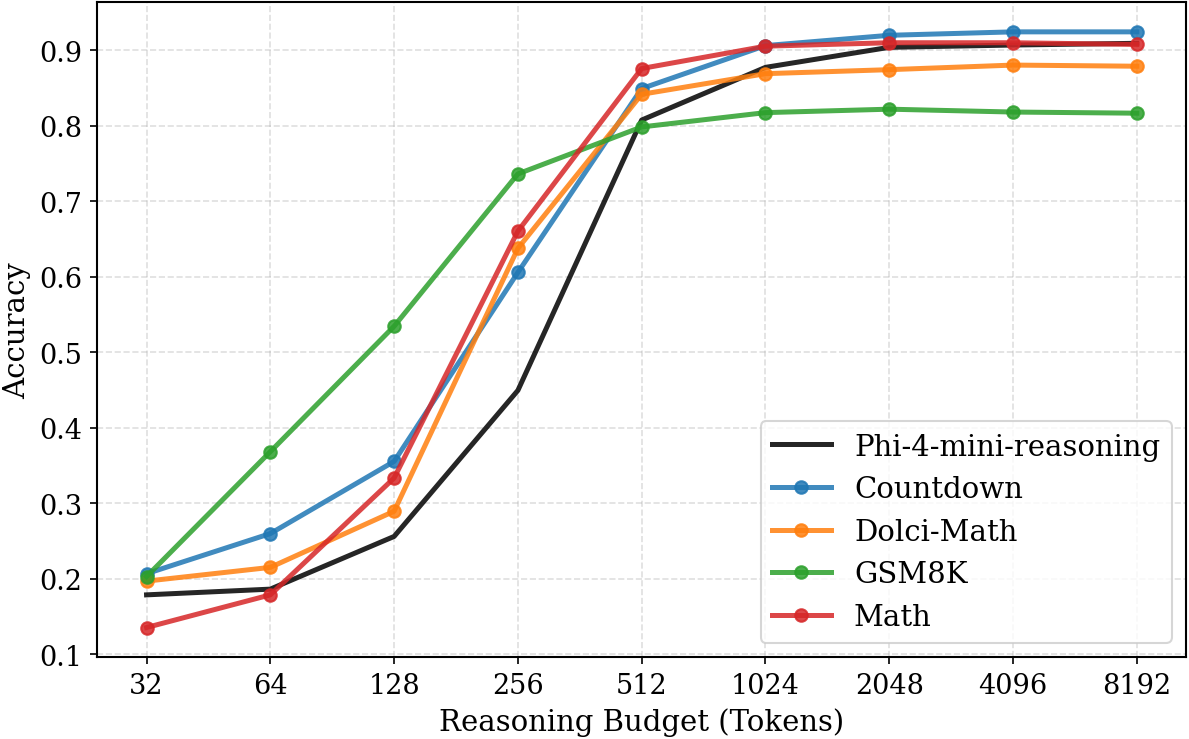}}
    \hfill
    \subfloat[Math500]{\includegraphics[width=0.44\linewidth]{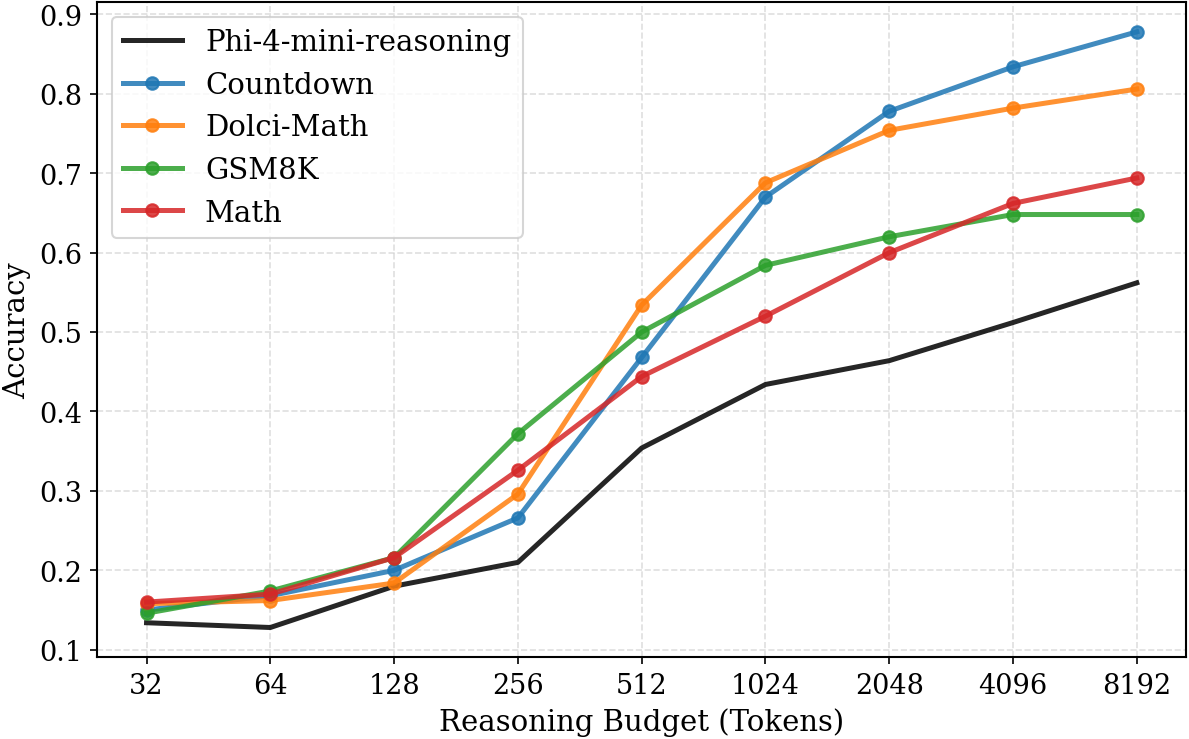}}
    \caption{Per-dataset accuracy as a function of the reasoning budget for Phi-4-mini-reasoning. The evaluation dataset is specified below the respective plot.}
    \label{fig:perdata_phi4mini}
    \vspace{-3mm}
\end{figure*}

We next evaluate whether the robustness learned through TRSD transfers to unseen prompt distributions. Specifically, we evaluate models on reasoning benchmarks that are not used during self-distillation, assessing whether robustness to truncated reasoning generalizes beyond the datasets seen during TRSD. For each architecture, we select the best-performing model across the evaluation datasets, as described in Appendix \ref{app:ckpt}. Results are summarized in Table \ref{tab:ood_combined}, with per-dataset performance curves reported in Figure \ref{fig:perdata_qwen3_4b} for Qwen3-4B and in Figure \ref{fig:perdata_phi4mini} for Phi-4-mini-reasoning.

Overall, TRSD generalizes well across prompt distributions. Under truncated inference, TRSD-trained models consistently outperform the corresponding baselines on out-of-distribution benchmarks, and in several cases match or exceed in-distribution performance. Performance gains are most pronounced in the low-budget regime and remain stable across datasets, even when the evaluation task differs from the self-distillation source. At larger reasoning budgets, performance typically approaches that of the baseline, with only trivial and task-dependent differences, mirroring the behavior observed in the in-distribution setting. Together, these results indicate that TRSD encourages a general ability to recover correct predictions from truncated reasoning, rather than adapting models to dataset-specific reasoning patterns.
\begin{table*}[t]
\centering
\caption{
Baseline vs TRSD performance under truncated inference on out-of-distribution benchmarks.
Entries report Baseline/TRSD accuracy (higher is better), bold indicates the better entry.
When Baseline and TRSD accuracies differ by at most 1 percentage point, both are bold.
Left: Baseline Qwen3-4B compared to TRSD trained on GSM8K.
Right: Baseline Phi-4-mini-reasoning compared to TRSD trained on Countdown.
}
\label{tab:ood_combined}

\scriptsize
\setlength{\tabcolsep}{2.6pt}
\renewcommand{\arraystretch}{1.05}

\begin{minipage}[t]{0.48\textwidth}
\centering
\resizebox{\linewidth}{!}{%
\begin{tabular}{lcccccc}
\toprule
\multicolumn{7}{c}{Teacher Model: Qwen3-4B, TRSD: GSM8K} \\
\midrule
\multirow{2}{*}{Budget / Data} &
\multicolumn{2}{c}{Countdown} &
\multicolumn{2}{c}{Dolci} &
\multicolumn{2}{c}{Math500} \\
\cmidrule(lr){2-3}\cmidrule(lr){4-5}\cmidrule(lr){6-7}
& Baseline & TRSD & Baseline & TRSD & Baseline & TRSD \\
\midrule
32   & \textbf{11.72} & \textbf{12.62} \textcolor{LimeGreen}{\scriptsize(+0.9)}
     & 8.85 & \textbf{12.18} \textcolor{LimeGreen}{\scriptsize(+3.3)}
     & 19.00 & \textbf{23.20} \textcolor{LimeGreen}{\scriptsize(+4.2)} \\
64   & 13.46 & \textbf{15.50} \textcolor{LimeGreen}{\scriptsize(+2.0)}
     & 8.97 & \textbf{12.14} \textcolor{LimeGreen}{\scriptsize(+3.2)}
     & 18.60 & \textbf{24.40} \textcolor{LimeGreen}{\scriptsize(+5.8)} \\
128  & \textbf{26.04} & 24.88 \textcolor{red}{\scriptsize(-1.2)}
     & 9.33 & \textbf{12.10} \textcolor{LimeGreen}{\scriptsize(+2.8)}
     & 21.20 & \textbf{27.40} \textcolor{LimeGreen}{\scriptsize(+6.2)} \\
256  & \textbf{43.94} & \textbf{44.36} \textcolor{LimeGreen}{\scriptsize(+0.4)}
     & 9.94 & \textbf{13.46} \textcolor{LimeGreen}{\scriptsize(+3.5)}
     & 26.00 & \textbf{39.00} \textcolor{LimeGreen}{\scriptsize(+13.0)} \\
512  & \textbf{60.28} & \textbf{60.82} \textcolor{LimeGreen}{\scriptsize(+0.5)}
     & 11.46 & \textbf{14.78} \textcolor{LimeGreen}{\scriptsize(+3.3)}
     & 45.40 & \textbf{58.80} \textcolor{LimeGreen}{\scriptsize(+13.4)} \\
1024 & \textbf{71.14} & \textbf{70.54} \textcolor{red}{\scriptsize(-0.6)}
     & 13.22 & \textbf{17.95} \textcolor{LimeGreen}{\scriptsize(+4.7)}
     & 67.00 & \textbf{74.00} \textcolor{LimeGreen}{\scriptsize(+7.0)} \\
2048 & \textbf{76.64} & \textbf{77.20} \textcolor{LimeGreen}{\scriptsize(+0.6)}
     & 17.19 & \textbf{21.55} \textcolor{LimeGreen}{\scriptsize(+4.4)}
     & \textbf{80.80} & \textbf{80.60} \textcolor{red}{\scriptsize(-0.2)} \\
4096 & \textbf{81.32} & \textbf{81.44} \textcolor{LimeGreen}{\scriptsize(+0.1)}
     & 23.32 & \textbf{29.13} \textcolor{LimeGreen}{\scriptsize(+5.8)}
     & \textbf{88.00} & 84.20 \textcolor{red}{\scriptsize(-3.8)} \\
8192 & \textbf{84.88} & 82.86 \textcolor{red}{\scriptsize(-2.0)}
     & 34.21 & \textbf{39.34} \textcolor{LimeGreen}{\scriptsize(+5.1)}
     & \textbf{91.40} & 86.20 \textcolor{red}{\scriptsize(-5.2)} \\
\bottomrule
\end{tabular}
}
\end{minipage}
\hfill
\begin{minipage}[t]{0.48\textwidth}
\centering
\resizebox{\linewidth}{!}{%
\begin{tabular}{lcccccc}
\toprule
\multicolumn{7}{c}{Teacher Model: Phi-4-mini-reasoning, TRSD: Countdown} \\
\midrule
\multirow{2}{*}{Budget / Data} &
\multicolumn{2}{c}{Dolci} &
\multicolumn{2}{c}{GSM8K} &
\multicolumn{2}{c}{Math500} \\
\cmidrule(lr){2-3}\cmidrule(lr){4-5}\cmidrule(lr){6-7}
& Baseline & TRSD & Baseline & TRSD & Baseline & TRSD \\
\midrule
32   & 7.05 & \textbf{9.54} \textcolor{LimeGreen}{\scriptsize(+2.5)}
     & 17.89 & \textbf{20.70} \textcolor{LimeGreen}{\scriptsize(+2.8)}
     & 13.40 & \textbf{15.00} \textcolor{LimeGreen}{\scriptsize(+1.6)} \\
64   & 6.73 & \textbf{9.98} \textcolor{LimeGreen}{\scriptsize(+3.3)}
     & 18.65 & \textbf{26.00} \textcolor{LimeGreen}{\scriptsize(+7.4)}
     & 12.80 & \textbf{16.80} \textcolor{LimeGreen}{\scriptsize(+4.0)} \\
128  & 8.33 & \textbf{10.10} \textcolor{LimeGreen}{\scriptsize(+1.8)}
     & 25.63 & \textbf{35.56} \textcolor{LimeGreen}{\scriptsize(+9.9)}
     & 18.00 & \textbf{20.00} \textcolor{LimeGreen}{\scriptsize(+2.0)} \\
256  & 9.58 & \textbf{10.58} \textcolor{LimeGreen}{\scriptsize(+1.0)}
     & 44.96 & \textbf{60.58} \textcolor{LimeGreen}{\scriptsize(+15.6)}
     & 21.00 & \textbf{26.60} \textcolor{LimeGreen}{\scriptsize(+5.6)} \\
512  & 9.62 & \textbf{11.70} \textcolor{LimeGreen}{\scriptsize(+2.1)}
     & 80.74 & \textbf{84.91} \textcolor{LimeGreen}{\scriptsize(+4.2)}
     & 35.40 & \textbf{46.80} \textcolor{LimeGreen}{\scriptsize(+11.4)} \\
1024 & 10.58 & \textbf{14.02} \textcolor{LimeGreen}{\scriptsize(+3.4)}
     & 87.72 & \textbf{90.60} \textcolor{LimeGreen}{\scriptsize(+2.9)}
     & 43.40 & \textbf{67.00} \textcolor{LimeGreen}{\scriptsize(+23.6)} \\
2048 & 13.50 & \textbf{16.23} \textcolor{LimeGreen}{\scriptsize(+2.7)}
     & 90.37 & \textbf{91.96} \textcolor{LimeGreen}{\scriptsize(+1.6)}
     & 46.40 & \textbf{77.80} \textcolor{LimeGreen}{\scriptsize(+31.4)} \\
4096 & 16.47 & \textbf{19.03} \textcolor{LimeGreen}{\scriptsize(+2.6)}
     & 90.67 & \textbf{92.42} \textcolor{LimeGreen}{\scriptsize(+1.8)}
     & 51.20 & \textbf{83.40} \textcolor{LimeGreen}{\scriptsize(+32.2)} \\
8192 & \textbf{24.44} & \textbf{24.24} \textcolor{red}{\scriptsize(-0.2)}
     & 90.90 & \textbf{92.42} \textcolor{LimeGreen}{\scriptsize(+1.5)}
     & 56.20 & \textbf{87.80} \textcolor{LimeGreen}{\scriptsize(+31.6)} \\
\bottomrule
\end{tabular}
}
\end{minipage}

\end{table*}

\subsection{Emergent Reduction in Reasoning Length}
\label{sec:emergent_shorter_reasoning}

While TRSD does not impose any explicit constraint on the length of the reasoning trace produced at inference time, its training objective discourages reliance on late-stage reasoning tokens that may be absent under truncation. As a result, TRSD-trained models consistently exhibit shorter reasoning traces even when inference is unconstrained.

We analyze this effect in detail by reporting the average number of reasoning tokens generated at the maximum reasoning budget ($r_{\max}=8192$), conditioned on whether the final prediction is correct or incorrect, in Table \ref{tab:thinklen_correct_wrong_combined}. TRSD consistently reduces reasoning length for both correct and incorrect predictions. This behavior is observed both for in-distribution evaluation and out-of-distribution. The reduction in reasoning length is particularly pronounced for Qwen3-4B. For Phi-4-mini-reasoning under out-of-distribution evaluation, reasoning length on GSM8K and MATH is comparable to, or slightly longer than, the baseline; however, this is accompanied by improved performance on both datasets.

\begin{table*}[t]
\centering

\caption{
Average number of reasoning tokens conditioned on answer correctness under unconstrained inference ($r_{\max}=8192$).
Top block: in-distribution TRSD (trained and evaluated on the same prompt distribution).
Bottom block: out-of-distribution TRSD using a single fixed checkpoint per model (Qwen3-4B trained on GSM8K; Phi-4-mini-reasoning trained on Countdown).
Bold indicates the configuration (Baseline or TRSD) with lower average reasoning length.
}
\label{tab:thinklen_correct_wrong_combined}
\resizebox{\columnwidth}{!}{
\begin{tabular}{@{}llcccccccc@{}}
\toprule
& & \multicolumn{2}{c}{Countdown} & \multicolumn{2}{c}{Dolci} & \multicolumn{2}{c}{GSM8K} & \multicolumn{2}{c}{Math500} \\
\cmidrule(lr){3-4}\cmidrule(lr){5-6}\cmidrule(lr){7-8}\cmidrule(lr){9-10}
Model & Outcome
& Baseline & TRSD
& Baseline & TRSD
& Baseline & TRSD
& Baseline & TRSD \\
\midrule

\multicolumn{10}{c}{\textit{In-distribution evaluation}} \\
\addlinespace[0.2em]
\midrule

\multirow{2}{*}{Qwen3-4B}
& Correct
& 1209 & \textbf{660}
& 6915 & \textbf{3328}
& 783  & \textbf{247}
& 2810 & \textbf{1570} \\
& Wrong
& 7169 & \textbf{3725}
& 7913 & \textbf{4458}
& 2546 & \textbf{661}
& 5770 & \textbf{5243} \\
\addlinespace[0.15em]
\midrule
\addlinespace[0.15em]

\multirow{2}{*}{Phi-4-mini-reasoning}
& Correct
& 1350 & \textbf{1095}
& 6272 & \textbf{4346}
& 683  & \textbf{513}
& 1859 & \textbf{1831} \\
& Wrong
& 4418 & \textbf{3799}
& 7585 & \textbf{6477}
& 1983 & \textbf{1537}
& 3523 & \textbf{3079} \\

\addlinespace[0.45em]
\cmidrule(lr){1-10}
\addlinespace[0.25em]

\multicolumn{10}{c}{\textit{Out-of-distribution evaluation (fixed TRSD checkpoint)}} \\
\addlinespace[0.2em]
\midrule

\multirow{2}{*}{Qwen3-4B}
& Correct
& 1209 & \textbf{824}
& 6915 & \textbf{5624}
& 783  & \textbf{247}
& 2810 & \textbf{1216} \\
& Wrong
& 7169 & \textbf{6307}
& 7913 & \textbf{7141}
& 2546 & \textbf{661}
& 5770 & \textbf{3135} \\
\addlinespace[0.15em]
\midrule
\addlinespace[0.15em]

\multirow{2}{*}{Phi-4-mini-reasoning}
& Correct
& 1350 & \textbf{1064}
& 6272 & \textbf{5625}
& 683  & \textbf{680}
& \textbf{1859} & 1893 \\
& Wrong
& 4418 & \textbf{3602}
& 7585 & \textbf{7484}
& \textbf{1983} & 2024
& \textbf{3523} & 5957 \\
\bottomrule
\end{tabular}
}
\vspace{-3mm}
\end{table*}

\subsection{Qualitative Analysis of Reasoning Traces}
\label{sec:qualitative}
Beyond aggregate accuracy and token-level statistics, we examine how TRSD affects the structure and length of model reasoning. We qualitatively compare reasoning traces produced by baseline and TRSD-trained models on representative examples where both models produce the correct answer. Figure~\ref{fig:qualitative_gsm8k} shows an example from GSM8K using Qwen3-4B. While both models arrive at the same final answer, the baseline produces a longer reasoning trace that repeatedly restates intermediate quantities and re-verifies simple arithmetic. In contrast, the TRSD-trained model reaches the solution using a more compact reasoning prefix that focuses on the essential computations. We provide more examples in Appendix \ref{app:qualitative} evidencing this same pattern: TRSD prunes unnecessary tokens and computation from a model's natural thinking trace, complementing the reasoning paradigm by successfully tackling its canonical inefficiencies.

\begin{figure*}[!t]
\centering
\small
\begin{tcolorbox}[
    colback=gray!3,
    colframe=gray!50,
    boxrule=0.5pt,
    arc=2pt,
    title={Qualitative comparison of reasoning traces (Qwen3-4B, GSM8K)}
]

\textbf{Prompt.}
Ray had 25 lollipops. He kept 5 lollipops and shared the remaining equally with his four friends.
How many lollipops did each of his friends receive?

\vspace{0.6em}

\begin{tabular}{p{0.46\linewidth} p{0.46\linewidth}}
\textbf{Baseline (Qwen3-4B)} &
\textbf{TRSD (GSM8K)} \\[0.4em]

\begin{minipage}[t]{\linewidth}
{\scriptsize
\textbf{Think:}\\
Okay, let's see. Ray had 25 lollipops. He kept 5 for himself, so the remaining ones are shared
equally with his four friends. Hmm, first I need to figure out how many lollipops were left after
he kept 5. So 25 total minus 5 he kept equals 20. Then he shares those 20 equally with his four
friends. So I need to divide 20 by 4. Let me check that again. 25 minus 5 is 20, right? Then
20 divided by 4 is 5. So each friend gets 5 lollipops. Wait, that seems straightforward. Did I
miss anything? The problem says ``shared the remaining equally with his four friends.'' So yeah,
20 divided by 4 is definitely 5. I think that's the answer.

\textbf{Answer:} 5
}
\end{minipage}
&
\begin{minipage}[t]{\linewidth}
\textbf{Think:}\\
Okay, let's see. Ray starts with 25 lollipops. He keeps 5 for himself, so we subtract those first.
That leaves 20 lollipops to share with his four friends. To find out how many each friend gets,
I divide 20 by 4. 20 divided by 4 is 5. So each friend gets 5 lollipops.

\textbf{Answer:} 5
\end{minipage}
\end{tabular}

\end{tcolorbox}

\vspace{-0.8em}
\caption{Example where both models answer correctly, but the TRSD-trained model uses a substantially
shorter reasoning trace. The example is taken verbatim from the Qwen3-4B GSM8K evaluation set.}
\label{fig:qualitative_gsm8k}
\vspace{-3mm}
\end{figure*}

\section{Related Work}
\label{sec5:related}
A common approach to reducing the cost of chain-of-thought reasoning is to explicitly shorten or prune reasoning traces. TokenSkip~\citep{xia2025tokenskip} skips intermediate reasoning tokens during generation based on learned importance scores, while Conditional Token Selection~\citep{yuan2025not} trains models to selectively retain only a subset of reasoning tokens. TokenSqueeze~\citep{zhang2025tokensqueeze} compresses reasoning traces using length-aware training objectives that encourage more compact outputs. Related work, such as O1-Pruner~\citep{luo2025o1} and CoT-Valve~\citep{ma2025cot}, incorporates length control directly into training, encouraging models to generate shorter reasoning traces through budget constraints or controllable reasoning length. Prompting-based methods such as Chain of Draft~\citep{xu2502chain} pursue a similar goal by instructing models to produce minimal intermediate reasoning at inference time. All these approaches explicitly optimize reasoning length or structure, either by deciding which tokens to keep, compress, or suppress, or by enforcing a target reasoning budget. In contrast, we do not enforce brevity, modify decoding, or alter the reasoning process itself. Instead, we train models to remain accurate when reasoning is arbitrarily truncated, allowing inference to stop early without explicitly selecting or pruning reasoning tokens. Any reduction in reasoning length observed at inference time emerges as a consequence of optimizing robustness to partial reasoning, rather than as an explicit training objective.

The work of \citet{chen2025distilling} analyzes which parts of reasoning traces provide effective supervision during distillation, showing that supervising selected portions of a teacher’s reasoning can retain strong performance. Adaptive Prefix Alignment~\citep{liu2026long} further explores how prefixes of reasoning traces can be used during distillation to reduce noise from later reasoning steps. These approaches typically distill from a larger teacher into a smaller student and rely on carefully selecting, weighting, or aligning specific parts of the teacher’s reasoning trace. In contrast, TRSD uses self-distillation, where the teacher and student share the same architecture and initialization, and does not assume that any particular portion of the reasoning is inherently more informative. By randomly truncating reasoning prefixes across the full range of possible truncation points, we directly optimize robustness of the answer distribution to partial reasoning, rather than optimizing for a fixed cutoff, compressed trace, or predefined notion of “important” reasoning tokens.
\section{Discussion and Future Work}
\label{sec6:discussion_conclusion}

In this section, we summarize the main empirical findings of TRSD, discuss its limitations, and outline directions for future work.

Across all evaluated architectures and datasets, TRSD consistently improves robustness to truncated inference, with the largest gains observed in low-budget regimes where baseline models frequently fail to complete a useful reasoning trajectory. At the same time, we note that the observed magnitude of the gains can be sensitive to the interaction between model architecture and the dataset used to generate self-distillation prompts. Different models achieve their largest improvements when trained on different prompt distributions, and no single dataset is uniformly optimal across architectures. While we do not identify a single underlying cause for this behavior, a plausible explanation is that models respond differently to prompt-induced reasoning patterns depending on their prior training, which shapes how reasoning traces are generated.

\subsection{Limitations and Failure Cases}

While TRSD consistently improves robustness under truncated inference, its gains naturally diminish as the available reasoning budget increases. 
In addition, improvements obtained through TRSD do not always translate uniformly across out-of-distribution prompt distributions. This sensitivity reflects the fact that TRSD optimizes robustness with respect to the reasoning patterns induced by the self-distillation prompts, and that different prompt distributions emphasize different reasoning structures. As a result, identifying suitable self-distillation prompts is an important factor in maximizing the benefits of the TRSD step, which we believe this work has only begun to explore.

\subsection{Future Directions}

Several directions could further improve the effectiveness and applicability of TRSD. First, alternative training strategies could be explored to mitigate the occasional performance inconsistencies observed at large reasoning budgets, for example, by adapting the distillation objective based on truncation level or teacher confidence. Such approaches may help better balance robustness to partial reasoning with performance under full-length inference. Second, our results indicate that the choice of self-distillation prompts influences downstream performance. Developing curated prompt datasets that elicit reasoning patterns that transfer reliably across tasks could improve the consistency of TRSD gains, particularly in out-of-distribution settings. Finally, while TRSD is applied here as a post-training step, robustness to truncated reasoning could be incorporated directly into the distillation process from larger teacher models, allowing robustness to partial reasoning to be learned jointly with standard distillation objectives rather than added as a separate training stage.

Overall, this work shows that robustness to truncated reasoning can be learned as a lightweight post-training property of existing reasoning models through self-distillation alone. By improving accuracy under limited reasoning budgets without introducing additional supervision or architectural changes, TRSD provides a practical approach for deploying reasoning models in settings where inference-time computation is constrained.

\bibliography{main}

@article{leviathan2025prompt,
  title={Prompt Repetition Improves Non-Reasoning LLMs},
  author={Leviathan, Yaniv and Kalman, Matan and Matias, Yossi},
  journal={arXiv preprint arXiv:2512.14982},
  year={2025}
}

@article{pipis2025wait,
  title={Wait, Wait, Wait... Why Do Reasoning Models Loop?},
  author={Pipis, Charilaos and Garg, Shivam and Kontonis, Vasilis and Shrivastava, Vaishnavi and Krishnamurthy, Akshay and Papailiopoulos, Dimitris},
  journal={arXiv preprint arXiv:2512.12895},
  year={2025}
}

@inproceedings{
chen2025do,
title={Do {NOT} Think That Much for 2+3=? On the Overthinking of Long Reasoning Models},
author={Xingyu Chen and Jiahao Xu and Tian Liang and Zhiwei He and Jianhui Pang and Dian Yu and Linfeng Song and Qiuzhi Liu and Mengfei Zhou and Zhuosheng Zhang and Rui Wang and Zhaopeng Tu and Haitao Mi and Dong Yu},
booktitle={Forty-second International Conference on Machine Learning},
year={2025},
url={https://openreview.net/forum?id=MSbU3L7V00}
}

@article{wang2025wait,
  title={Wait, We Don't Need to" Wait"! Removing Thinking Tokens Improves Reasoning Efficiency},
  author={Wang, Chenlong and Feng, Yuanning and Chen, Dongping and Chu, Zhaoyang and Krishna, Ranjay and Zhou, Tianyi},
  journal={arXiv preprint arXiv:2506.08343},
  year={2025}
}

@inproceedings{yan2025long,
  title={From Long to Lean: Performance-aware and Adaptive Chain-of-Thought Compression via Multi-round Refinement},
  author={Yan, Jianzhi and Liu, Le and Pan, Youcheng and Chen, Shiwei and Yuan, Zike and Xiang, Yang and Tang, Buzhou},
  booktitle={Proceedings of the 2025 Conference on Empirical Methods in Natural Language Processing},
  pages={12290--12306},
  year={2025}
}

@article{zhang2023fast,
  title={Fast chain-of-thought: A glance of future from parallel decoding leads to answers faster},
  author={Zhang, Hongxuan and Liu, Zhining and Zhao, Yao and Zheng, Jiaqi and Zhuang, Chenyi and Gu, Jinjie and Chen, Guihai},
  journal={arXiv preprint arXiv:2311.08263},
  year={2023}
}

@article{xia2025tokenskip,
  title={Tokenskip: Controllable chain-of-thought compression in llms},
  author={Xia, Heming and Leong, Chak Tou and Wang, Wenjie and Li, Yongqi and Li, Wenjie},
  journal={arXiv preprint arXiv:2502.12067},
  year={2025}
}

@inproceedings{
zhang2025tokensqueeze,
title={TokenSqueeze: Performance-Preserving Compression for Reasoning {LLM}s},
author={Yuxiang Zhang and Zhengxu Yu and Weihang Pan and Zhongming Jin and Qiang Fu and Deng Cai and Binbin Lin and Jieping Ye},
booktitle={The Thirty-ninth Annual Conference on Neural Information Processing Systems},
year={2025},
url={https://openreview.net/forum?id=Wc1VZ2bVJn}
}

@article{zhang2025lightthinker,
  title={Lightthinker: Thinking step-by-step compression},
  author={Zhang, Jintian and Zhu, Yuqi and Sun, Mengshu and Luo, Yujie and Qiao, Shuofei and Du, Lun and Zheng, Da and Chen, Huajun and Zhang, Ningyu},
  journal={arXiv preprint arXiv:2502.15589},
  year={2025}
}

@inproceedings{muennighoff2025s1,
  title={s1: Simple test-time scaling},
  author={Muennighoff, Niklas and Yang, Zitong and Shi, Weijia and Li, Xiang Lisa and Fei-Fei, Li and Hajishirzi, Hannaneh and Zettlemoyer, Luke and Liang, Percy and Cand{\`e}s, Emmanuel and Hashimoto, Tatsunori B},
  booktitle={Proceedings of the 2025 Conference on Empirical Methods in Natural Language Processing},
  pages={20286--20332},
  year={2025}
}

@article{wei2022chain,
  title={Chain-of-thought prompting elicits reasoning in large language models},
  author={Wei, Jason and Wang, Xuezhi and Schuurmans, Dale and Bosma, Maarten and Xia, Fei and Chi, Ed and Le, Quoc V and Zhou, Denny and others},
  journal={Advances in neural information processing systems},
  volume={35},
  pages={24824--24837},
  year={2022}
}

@article{guo2025deepseek,
  title={Deepseek-r1: Incentivizing reasoning capability in llms via reinforcement learning},
  author={Guo, Daya and Yang, Dejian and Zhang, Haowei and Song, Junxiao and Zhang, Ruoyu and Xu, Runxin and Zhu, Qihao and Ma, Shirong and Wang, Peiyi and Bi, Xiao and others},
  journal={arXiv preprint arXiv:2501.12948},
  year={2025}
}

@article{yang2025qwen3,
  title={Qwen3 technical report},
  author={Yang, An and Li, Anfeng and Yang, Baosong and Zhang, Beichen and Hui, Binyuan and Zheng, Bo and Yu, Bowen and Gao, Chang and Huang, Chengen and Lv, Chenxu and others},
  journal={arXiv preprint arXiv:2505.09388},
  year={2025}
}

@article{chen2025distilling,
  title={Distilling the Essence: Efficient Reasoning Distillation via Sequence Truncation},
  author={Chen, Wei-Rui and Kothapalli, Vignesh and Fatahibaarzi, Ata and Sang, Hejian and Tang, Shao and Song, Qingquan and Wang, Zhipeng and Abdul-Mageed, Muhammad},
  journal={arXiv preprint arXiv:2512.21002},
  year={2025}
}

@article{ma2025cot,
  title={Cot-valve: Length-compressible chain-of-thought tuning, 2025},
  author={Ma, Xinyin and Wan, Guangnian and Yu, Runpeng and Fang, Gongfan and Wang, Xinchao},
  journal={URL https://arxiv. org/abs/2502},
  volume={9601},
  year={2025}
}

@article{xu2502chain,
  title={Chain of draft: Thinking faster by writing less, 2025a},
  author={Xu, Silei and Xie, Wenhao and Zhao, Lingxiao and He, Pengcheng},
  journal={URL https://arxiv. org/abs/2502.18600}
}

@article{luo2025o1,
  title={O1-pruner: Length-harmonizing fine-tuning for o1-like reasoning pruning},
  author={Luo, Haotian and Shen, Li and He, Haiying and Wang, Yibo and Liu, Shiwei and Li, Wei and Tan, Naiqiang and Cao, Xiaochun and Tao, Dacheng},
  journal={arXiv preprint arXiv:2501.12570},
  year={2025}
}

@article{yuan2025not,
  title={Not All Tokens Are What You Need In Thinking},
  author={Yuan, Hang and Yu, Bin and Li, Haotian and Yang, Shijun and Wang, Christina Dan and Yu, Zhou and Xu, Xueyin and Qi, Weizhen and Chen, Kai},
  journal={arXiv preprint arXiv:2505.17827},
  year={2025}
}

@article{guha2025openthoughts,
  title={OpenThoughts: Data Recipes for Reasoning Models},
  author={Guha, Etash and Marten, Ryan and Keh, Sedrick and Raoof, Negin and Smyrnis, Georgios and Bansal, Hritik and Nezhurina, Marianna and Mercat, Jean and Vu, Trung and Sprague, Zayne and others},
  journal={arXiv preprint arXiv:2506.04178},
  year={2025}
}

@article{ye2025limo,
  title={Limo: Less is more for reasoning},
  author={Ye, Yixin and Huang, Zhen and Xiao, Yang and Chern, Ethan and Xia, Shijie and Liu, Pengfei},
  journal={arXiv preprint arXiv:2502.03387},
  year={2025}
}

@article{liu2026long,
  title={Long-Chain Reasoning Distillation via Adaptive Prefix Alignment},
  author={Liu, Zhenghao and Wu, Zhuoyang and Li, Xinze and Yan, Yukun and Wang, Shuo and Chen, Zulong and Gu, Yu and Yu, Ge and Sun, Maosong},
  journal={arXiv preprint arXiv:2601.10064},
  year={2026}
}

@article{xu2025phi,
  title={Phi-4-mini-reasoning: Exploring the limits of small reasoning language models in math},
  author={Xu, Haoran and Peng, Baolin and Awadalla, Hany and Chen, Dongdong and Chen, Yen-Chun and Gao, Mei and Kim, Young Jin and Li, Yunsheng and Ren, Liliang and Shen, Yelong and others},
  journal={arXiv preprint arXiv:2504.21233},
  year={2025}
}

@article{cobbe2021gsm8k,
  title={Training Verifiers to Solve Math Word Problems},
  author={Cobbe, Karl and Kosaraju, Vineet and Bavarian, Mohammad and Chen, Mark and Jun, Heewoo and Kaiser, Lukasz and Plappert, Matthias and Tworek, Jerry and Hilton, Jacob and Nakano, Reiichiro and Hesse, Christopher and Schulman, John},
  journal={arXiv preprint arXiv:2110.14168},
  year={2021}
}

@misc{olmo2025olmo3,
title={Olmo 3},
author={Team Olmo and Allyson Ettinger and Amanda Bertsch and Bailey Kuehl and David Graham and David Heineman and Dirk Groeneveld and Faeze Brahman and Finbarr Timbers and Hamish Ivison and Jacob Morrison and Jake Poznanski and Kyle Lo and Luca Soldaini and Matt Jordan and Mayee Chen and Michael Noukhovitch and Nathan Lambert and Pete Walsh and Pradeep Dasigi and Robert Berry and Saumya Malik and Saurabh Shah and Scott Geng and Shane Arora and Shashank Gupta and Taira Anderson and Teng Xiao and Tyler Murray and Tyler Romero and Victoria Graf and Akari Asai and Akshita Bhagia and Alexander Wettig and Alisa Liu and Aman Rangapur and Chloe Anastasiades and Costa Huang and Dustin Schwenk and Harsh Trivedi and Ian Magnusson and Jaron Lochner and Jiacheng Liu and Lester James V. Miranda and Maarten Sap and Malia Morgan and Michael Schmitz and Michal Guerquin and Michael Wilson and Regan Huff and Ronan Le Bras and Rui Xin and Rulin Shao and Sam Skjonsberg and Shannon Zejiang Shen and Shuyue Stella Li and Tucker Wilde and Valentina Pyatkin and Will Merrill and Yapei Chang and Yuling Gu and Zhiyuan Zeng and Ashish Sabharwal and Luke Zettlemoyer and Pang Wei Koh and Ali Farhadi and Noah A. Smith and Hannaneh Hajishirzi},
year={2025},
eprint={2512.13961},
archivePrefix={arXiv},
primaryClass={cs.CL},
url={https://arxiv.org/abs/2512.13961},
}

@article{hendrycksmath2021,
  title={Measuring Mathematical Problem Solving With the MATH Dataset},
  author={Dan Hendrycks and Collin Burns and Saurav Kadavath and Akul Arora and Steven Basart and Eric Tang and Dawn Song and Jacob Steinhardt},
  journal={NeurIPS},
  year={2021}
}
\bibliographystyle{icml2026}

\newpage
\appendix

\section{Implementation Details}
\label{appA:implementation}
\subsection{Teacher Answer Processing}
\label{app:teacher_answer}
When generating outputs, the teacher model does not always strictly follow the prescribed format, namely a reasoning trace enclosed in \texttt{<think></think>} tags followed by a final answer enclosed in \texttt{<answer></answer>} tags. In addition, generation may terminate early due to exhaustion of the reasoning token budget. We therefore apply a deterministic post-processing procedure to sanitize teacher outputs before they are used for distillation. First, if the generated output does not begin with a \texttt{<think>} tag, we prepend one. If a closing \texttt{</think>} tag is missing, we append it at the end of the generated text and mark the example for answer forcing. We then inspect the text following the final \texttt{</think>} tag. If no \texttt{<answer>} tag is present, we discard any trailing content and prompt the model to generate an answer by appending \texttt{<answer>} to the sanitized prefix, following the answer forcing procedure of \citet{muennighoff2025s1}. If an \texttt{<answer>} tag is present but the corresponding closing \texttt{</answer>} tag is missing, we similarly continue generation until the answer is completed. In all answer-forcing cases, we allocate a maximum budget of 200 tokens for answer generation. If multiple \texttt{<answer></answer>} pairs are present in the output, only the first pair is retained and all subsequent content is discarded. This procedure ensures that each training example contains a well-formed reasoning block and a single, clearly delimited answer segment.

\subsection{Training details}
\label{app:training_details}
All models are fine-tuned with TRSD for 2000 optimization steps using the AdamW optimizer, with learning rate $3\times 10^{-6}$, batch size 32, $\beta_1 = 0.9$, $\beta_2 = 0.999$, $\epsilon = 10^{-8}$, no weight decay, and gradient norm clipping at $1.0$. We save intermediate checkpoints at steps 250, 500, 1000, and 2000 and evaluate each checkpoint using the protocol described in Section \ref{app:ckpt}. We use temperature $1$ to generate reasoning and answers during training, and $0.8$ at evaluation.

\subsection{Checkpoint selection}
\label{app:ckpt}
We use different checkpoint selection rules for in-distribution and out-of-distribution evaluation.

\paragraph{In-distribution.}
For each model and each training dataset, we select the checkpoint that achieves the best performance on the corresponding evaluation set of that same dataset (i.e., the dataset used to draw prompts during self-distillation). These checkpoints are used for all in-distribution results.

\paragraph{Out-of-distribution.}
For each model, we select a single checkpoint to be used across all out-of-distribution evaluations. Concretely, among the four candidate checkpoints (steps 250/500/1000/2000), we select the checkpoint that performs best on average across all four evaluation datasets, including the in-distribution dataset. These checkpoints are used for all out-of-distribution results.

Table~\ref{tab:ckpt_selection} summarizes the checkpoints selected for in-distribution and out-of-distribution evaluation.

\begin{table}[ht]
\centering
\caption{Checkpoint selection summary. Columns corresponding to a specific dataset report the checkpoints achieving the best in-distribution performance for each (model, dataset) pair. The final column reports the checkpoint used for out-of-distribution evaluation, selected by best average performance across all four datasets; the dataset used for self-distillation is shown in parentheses. All entries denote the training step of the selected checkpoint.}
\begin{tabular}{lcccc|c}
\toprule
\textbf{Model} & \textbf{Countdown} & \textbf{Dolci} & \textbf{GSM8K} & \textbf{MATH500} & \textbf{OOD checkpoint} \\
\midrule
Qwen3-0.6B & 500 & 2000 & 250 & 250 & 500 (Countdown) \\
Qwen3-1.7B & 500 & 1000 & 250 & 250 & 250 (GSM8K) \\
Qwen3-4B   & 250 & 2000 & 250 & 250 & 250 (GSM8K) \\
Phi-4-mini-reasoning & 1000 & 1000 & 500 & 500 & 500 (Countdown) \\
OpenThinker3-1.5B & 1000 & 250 & 2000 & 1000 & 2000 (GSM8K) \\
\bottomrule
\end{tabular}
\label{tab:ckpt_selection}
\end{table}

\section{Datasets}
\label{appB:datasets}
We evaluate TRSD on four reasoning benchmarks spanning arithmetic search, grade-school math word problems, and competition-style mathematics. For all datasets, prompts are formatted to request reasoning inside \texttt{<think></think>} tags and a final answer inside \texttt{<answer></answer>} tags (see Appendix~\ref{app:teacher_answer} for how we sanitize teacher outputs when this format is not followed).

\paragraph{Countdown.}
We use a preconstructed version of the Countdown arithmetic-construction task from the \texttt{Jiayi-Pan/Countdown-Tasks-3to4} dataset on Hugging Face. Each example provides a multiset of integers and a target value, and the model must construct an equation that reaches the target using basic arithmetic operations while using each number at most once.
We subsample the dataset and split it into 45000 training prompts and 5000 test prompts.
A typical prompt is:
\begin{quote}\small
\texttt{Using the numbers 33, 5, 68, 29, create an equation that equals 67. You can use basic arithmetic operations (+, -, *, /) one or multiple times but each number can only be used once. Show your work in the <think> </think> tags and return the final equation in the <answer> </answer> tags, for example <answer> (1 + 2) / 3 </answer>. Think step by step inside <think> tags.}
\end{quote}

\paragraph{MATH.}
For competition-style mathematics, we use the MATH dataset of \citet{hendrycksmath2021} via the \texttt{EleutherAI/hendrycks\_math} release on Hugging Face.
We draw the 7500 training prompts from the official training split and evaluate on the standard MATH500 subset (500 problems), following common practice for efficient evaluation.
Prompts follow a standard instruction format:
\begin{quote}\small
\texttt{Solve the math problem. Think step-by-step inside <think> tags, then put only the final answer inside <answer> tags.
Problem:
The point $(a, b)$ lies on the line with the equation $3x + 2y = 12.$ When $a = 4$, what is the value of $b$?}
\end{quote}

\paragraph{GSM8K.}
We use GSM8K \citep{cobbe2021gsm8k} through the \texttt{openai/gsm8k} Hugging Face dataset.
GSM8K consists of grade-school math word problems requiring multi-step reasoning. The dataset is composed by 7470 training prompts and 1320 evaluation prompts.
A typical prompt is:
\begin{quote}\small
\texttt{Solve the math word problem. Think step-by-step inside <think> tags, then put only the final integer answer inside <answer> tags.
Question:
Darrell and Allen's ages are in the ratio of 7:11. If their total age now is 162, calculate Allen's age 10 years from now.}
\end{quote}

\paragraph{Dolci (math subset).}
We use the math portion of the \texttt{allenai/Dolci-Think-RL-7B} dataset, a collection of prompts designed to elicit deliberate reasoning \citep{olmo2025olmo3}.
We split the 24951 available examples into 22455 training and 2496 test prompts.
An example prompt is:
\begin{quote}\small
\texttt{user: What is the probability of such event happening: Form a word by randomly choosing 2 letters from the multiset {x: 3, l: 4}, shuffle the letters in the word, what is the probability of no letter 'x' occupy any of their original positions?
If the probability can be written as the form $\frac{m}{n}$, where $m$ and $n$ are relatively prime integers, find $m + n$.
Show your work in the <think> </think> tags and return the final equation in the <answer> </answer> tags.}
\end{quote}
\section{Extended Results}
\label{appD:extended_results}

We report here extended experimental results that complement the main findings in Section \ref{sec4:experiments}.
In particular, we provide full in-distribution and out-of-distribution tables for additional model sizes and architectures, together with per-dataset plots that visualize performance as a function of the available reasoning budget.
All results follow the same evaluation protocol and checkpoint selection strategy described in Section \ref{sec4:experiments} and Appendix \ref{app:ckpt}.

\subsection{Extended tables}
\label{appD:tables}

We first report extended numerical results for additional model families omitted from the main text.
In-distribution performance under truncated inference for Qwen3-0.6B, Qwen3-1.7B, and OpenThinker3-1.5B is shown in Table ~\ref{tab:small_models_in_combined}.

Across all three architectures, the same qualitative pattern observed in Section \ref{sec4:experiments} holds.
TRSD yields the largest gains in the heavily truncated regime, where only a small number of reasoning tokens are available and baseline models often fail to complete a useful reasoning trace.
As the reasoning budget increases, the performance gap typically narrows, and in some cases reverses slightly at the largest budgets.

Out-of-distribution generalization results for the same models are reported in Table \ref{tab:small_models_out_combined}.
Following the checkpoint selection procedure described in Appendix \ref{app:ckpt}, a single checkpoint per model is used for all out-of-distribution evaluations.
Despite this restriction, TRSD-trained models retain most of their gains under truncated inference, indicating that robustness learned from partial reasoning transfers across prompt distributions.
\begin{table*}[h]
\centering
\caption{
Baseline vs TRSD accuracy under truncated inference for smaller and mid-size models (in-distribution prompts).
Entries report Baseline/TRSD accuracy (higher is better). Colored values report the absolute change in accuracy
(TRSD $-$ Baseline, percentage points). Bold indicates the better entry; when Baseline and TRSD differ by at most
1 percentage point, both are bold.
}
\label{tab:small_models_in_combined}
\resizebox{\columnwidth}{!}{
\begin{tabular}{lcccccccc}
\toprule
\multirow{2}{*}{Budget / Data} &
\multicolumn{2}{c}{Countdown} &
\multicolumn{2}{c}{Dolci} &
\multicolumn{2}{c}{GSM8K} &
\multicolumn{2}{c}{Math500} \\
\cmidrule(lr{4pt}){2-3}
\cmidrule(lr{4pt}){4-5}
\cmidrule(lr{4pt}){6-7}
\cmidrule(l{6pt}){8-9}
& Baseline & TRSD &
Baseline & TRSD &
Baseline & TRSD &
Baseline & TRSD \\
\midrule

\multicolumn{9}{c}{\emph{Teacher Model: Qwen3-0.6B}} \\
\midrule
32   & 14.06 & \textbf{22.14} \textcolor{LimeGreen}{\scriptsize(\hphantom{-}+8.1)}
     & 3.37  & \textbf{4.45}  \textcolor{LimeGreen}{\scriptsize(\hphantom{-}+1.1)}
     & 3.03  & \textbf{8.34}  \textcolor{LimeGreen}{\scriptsize(\hphantom{-}+5.3)}
     & 7.20  & \textbf{11.20} \textcolor{LimeGreen}{\scriptsize(\hphantom{-}+4.0)} \\
64   & 13.84 & \textbf{22.12} \textcolor{LimeGreen}{\scriptsize(\hphantom{-}+8.3)}
     & \textbf{3.89} & \textbf{4.69} \textcolor{LimeGreen}{\scriptsize(\hphantom{-}+0.8)}
     & 3.41  & \textbf{10.24} \textcolor{LimeGreen}{\scriptsize(\hphantom{-}+6.8)}
     & 6.40  & \textbf{11.00} \textcolor{LimeGreen}{\scriptsize(\hphantom{-}+4.6)} \\
128  & 17.18 & \textbf{29.86} \textcolor{LimeGreen}{\scriptsize(\hphantom{-}+12.7)}
     & 3.37  & \textbf{5.17}  \textcolor{LimeGreen}{\scriptsize(\hphantom{-}+1.8)}
     & 4.47  & \textbf{25.85} \textcolor{LimeGreen}{\scriptsize(\hphantom{-}+21.4)}
     & 7.40  & \textbf{11.60} \textcolor{LimeGreen}{\scriptsize(\hphantom{-}+4.2)} \\
256  & 25.82 & \textbf{42.46} \textcolor{LimeGreen}{\scriptsize(\hphantom{-}+16.6)}
     & 3.93  & \textbf{5.57}  \textcolor{LimeGreen}{\scriptsize(\hphantom{-}+1.6)}
     & 23.96 & \textbf{54.59} \textcolor{LimeGreen}{\scriptsize(\hphantom{-}+30.6)}
     & 12.40 & \textbf{22.60} \textcolor{LimeGreen}{\scriptsize(\hphantom{-}+10.2)} \\
512  & 40.24 & \textbf{53.56} \textcolor{LimeGreen}{\scriptsize(\hphantom{-}+13.3)}
     & \textbf{4.81} & \textbf{5.69} \textcolor{LimeGreen}{\scriptsize(\hphantom{-}+0.9)}
     & 59.14 & \textbf{67.55} \textcolor{LimeGreen}{\scriptsize(\hphantom{-}+8.4)}
     & 31.80 & \textbf{37.00} \textcolor{LimeGreen}{\scriptsize(\hphantom{-}+5.2)} \\
1024 & 52.54 & \textbf{61.22} \textcolor{LimeGreen}{\scriptsize(\hphantom{-}+8.7)}
     & \textbf{5.05} & \textbf{5.89} \textcolor{LimeGreen}{\scriptsize(\hphantom{-}+0.8)}
     & 70.81 & \textbf{72.71} \textcolor{LimeGreen}{\scriptsize(\hphantom{-}+1.9)}
     & 47.40 & \textbf{50.00} \textcolor{LimeGreen}{\scriptsize(\hphantom{-}+2.6)} \\
2048 & 61.76 & \textbf{66.54} \textcolor{LimeGreen}{\scriptsize(\hphantom{-}+4.8)}
     & \textbf{6.85} & \textbf{5.97} \textcolor{red}{\scriptsize(\hphantom{+}-0.9)}
     & \textbf{74.60} & \textbf{74.45} \textcolor{red}{\scriptsize(\hphantom{+}-0.2)}
     & \textbf{60.80} & 54.40 \textcolor{red}{\scriptsize(\hphantom{+}-6.4)} \\
4096 & \textbf{68.44} & \textbf{68.58} \textcolor{LimeGreen}{\scriptsize(\hphantom{-}+0.1)}
     & \textbf{7.29} & \textbf{6.29} \textcolor{red}{\scriptsize(\hphantom{+}-1.0)}
     & \textbf{76.35} & \textbf{75.36} \textcolor{red}{\scriptsize(\hphantom{+}-1.0)}
     & \textbf{67.00} & 55.60 \textcolor{red}{\scriptsize(\hphantom{+}-11.4)} \\
8192 & \textbf{70.30} & 68.74 \textcolor{red}{\scriptsize(\hphantom{+}-1.6)}
     & \textbf{6.77} & \textbf{6.13} \textcolor{red}{\scriptsize(\hphantom{+}-0.6)}
     & \textbf{76.50} & 74.91 \textcolor{red}{\scriptsize(\hphantom{+}-1.6)}
     & \textbf{70.40} & 56.60 \textcolor{red}{\scriptsize(\hphantom{+}-13.8)} \\

\midrule
\multicolumn{9}{c}{\emph{Teacher Model: Qwen3-1.7B}} \\
\midrule
32   & 10.68 & \textbf{25.98} \textcolor{LimeGreen}{\scriptsize(\hphantom{-}+15.3)}
     & 4.37  & \textbf{6.21}  \textcolor{LimeGreen}{\scriptsize(\hphantom{-}+1.8)}
     & 6.82  & \textbf{15.09} \textcolor{LimeGreen}{\scriptsize(\hphantom{-}+8.3)}
     & 11.60 & \textbf{14.80} \textcolor{LimeGreen}{\scriptsize(\hphantom{-}+3.2)} \\
64   & 13.50 & \textbf{26.70} \textcolor{LimeGreen}{\scriptsize(\hphantom{-}+13.2)}
     & 3.89  & \textbf{6.93}  \textcolor{LimeGreen}{\scriptsize(\hphantom{-}+3.0)}
     & 6.90  & \textbf{17.36} \textcolor{LimeGreen}{\scriptsize(\hphantom{-}+10.5)}
     & 11.20 & \textbf{16.40} \textcolor{LimeGreen}{\scriptsize(\hphantom{-}+5.2)} \\
128  & 19.40 & \textbf{30.20} \textcolor{LimeGreen}{\scriptsize(\hphantom{-}+10.8)}
     & 3.81  & \textbf{6.77}  \textcolor{LimeGreen}{\scriptsize(\hphantom{-}+3.0)}
     & 9.40  & \textbf{32.75} \textcolor{LimeGreen}{\scriptsize(\hphantom{-}+23.4)}
     & 12.20 & \textbf{20.20} \textcolor{LimeGreen}{\scriptsize(\hphantom{-}+8.0)} \\
256  & 39.24 & \textbf{42.94} \textcolor{LimeGreen}{\scriptsize(\hphantom{-}+3.7)}
     & 4.29  & \textbf{6.81}  \textcolor{LimeGreen}{\scriptsize(\hphantom{-}+2.5)}
     & 33.06 & \textbf{65.88} \textcolor{LimeGreen}{\scriptsize(\hphantom{-}+32.8)}
     & 17.80 & \textbf{24.00} \textcolor{LimeGreen}{\scriptsize(\hphantom{-}+6.2)} \\
512  & 55.60 & \textbf{57.38} \textcolor{LimeGreen}{\scriptsize(\hphantom{-}+1.8)}
     & \textbf{6.57} & \textbf{6.77} \textcolor{LimeGreen}{\scriptsize(\hphantom{-}+0.2)}
     & 69.67 & \textbf{81.65} \textcolor{LimeGreen}{\scriptsize(\hphantom{-}+12.0)}
     & 38.00 & \textbf{44.60} \textcolor{LimeGreen}{\scriptsize(\hphantom{-}+6.6)} \\
1024 & \textbf{66.10} & \textbf{66.42} \textcolor{LimeGreen}{\scriptsize(\hphantom{-}+0.3)}
     & \textbf{8.17} & \textbf{8.21} \textcolor{LimeGreen}{\scriptsize(\hphantom{-}+0.0)}
     & 84.53 & \textbf{87.72} \textcolor{LimeGreen}{\scriptsize(\hphantom{-}+3.2)}
     & 60.00 & \textbf{64.60} \textcolor{LimeGreen}{\scriptsize(\hphantom{-}+4.6)} \\
2048 & \textbf{72.60} & \textbf{71.76} \textcolor{red}{\scriptsize(\hphantom{+}-0.8)}
     & \textbf{9.82} & \textbf{9.78} \textcolor{red}{\scriptsize(\hphantom{+}-0.0)}
     & \textbf{88.25} & \textbf{88.70} \textcolor{LimeGreen}{\scriptsize(\hphantom{-}+0.5)}
     & \textbf{74.20} & \textbf{74.80} \textcolor{LimeGreen}{\scriptsize(\hphantom{-}+0.6)} \\
4096 & \textbf{77.54} & 73.28 \textcolor{red}{\scriptsize(\hphantom{+}-4.3)}
     & \textbf{12.78} & 10.66 \textcolor{red}{\scriptsize(\hphantom{+}-2.1)}
     & \textbf{89.61} & \textbf{88.86} \textcolor{red}{\scriptsize(\hphantom{+}-0.8)}
     & \textbf{80.40} & 79.20 \textcolor{red}{\scriptsize(\hphantom{+}-1.2)} \\
8192 & \textbf{80.16} & 73.32 \textcolor{red}{\scriptsize(\hphantom{+}-6.8)}
     & \textbf{17.23} & 12.26 \textcolor{red}{\scriptsize(\hphantom{+}-5.0)}
     & \textbf{89.92} & \textbf{89.16} \textcolor{red}{\scriptsize(\hphantom{+}-0.8)}
     & \textbf{84.60} & 82.20 \textcolor{red}{\scriptsize(\hphantom{+}-2.4)} \\

\midrule
\multicolumn{9}{c}{\emph{Teacher Model: OpenThinker3-1.5B}} \\
\midrule
32   & 0.26 & \textbf{12.96} \textcolor{LimeGreen}{\scriptsize(\hphantom{-}+12.7)}
     & 3.73 & \textbf{4.81}  \textcolor{LimeGreen}{\scriptsize(\hphantom{-}+1.1)}
     & 2.58 & \textbf{7.28}  \textcolor{LimeGreen}{\scriptsize(\hphantom{-}+4.7)}
     & 3.40 & \textbf{13.20} \textcolor{LimeGreen}{\scriptsize(\hphantom{-}+9.8)} \\
64   & 0.46 & \textbf{15.64} \textcolor{LimeGreen}{\scriptsize(\hphantom{-}+15.2)}
     & \textbf{4.49} & \textbf{4.89} \textcolor{LimeGreen}{\scriptsize(\hphantom{-}+0.4)}
     & 3.34 & \textbf{7.28}  \textcolor{LimeGreen}{\scriptsize(\hphantom{-}+3.9)}
     & 6.00 & \textbf{13.20} \textcolor{LimeGreen}{\scriptsize(\hphantom{-}+7.2)} \\
128  & 0.82 & \textbf{17.30} \textcolor{LimeGreen}{\scriptsize(\hphantom{-}+16.5)}
     & \textbf{4.37} & \textbf{4.77} \textcolor{LimeGreen}{\scriptsize(\hphantom{-}+0.4)}
     & 4.02 & \textbf{8.26}  \textcolor{LimeGreen}{\scriptsize(\hphantom{-}+4.2)}
     & 6.20 & \textbf{15.00} \textcolor{LimeGreen}{\scriptsize(\hphantom{-}+8.8)} \\
256  & 4.84 & \textbf{26.36} \textcolor{LimeGreen}{\scriptsize(\hphantom{-}+21.5)}
     & 3.37 & \textbf{5.49}  \textcolor{LimeGreen}{\scriptsize(\hphantom{-}+2.1)}
     & 9.33 & \textbf{22.59} \textcolor{LimeGreen}{\scriptsize(\hphantom{-}+13.3)}
     & 11.20 & \textbf{19.00} \textcolor{LimeGreen}{\scriptsize(\hphantom{-}+7.8)} \\
512  & 16.64 & \textbf{36.08} \textcolor{LimeGreen}{\scriptsize(\hphantom{-}+19.4)}
     & 4.65 & \textbf{5.93}  \textcolor{LimeGreen}{\scriptsize(\hphantom{-}+1.3)}
     & 35.25 & \textbf{54.81} \textcolor{LimeGreen}{\scriptsize(\hphantom{-}+19.6)}
     & 24.80 & \textbf{29.00} \textcolor{LimeGreen}{\scriptsize(\hphantom{-}+4.2)} \\
1024 & 28.28 & \textbf{44.42} \textcolor{LimeGreen}{\scriptsize(\hphantom{-}+16.1)}
     & 4.53 & \textbf{7.21}  \textcolor{LimeGreen}{\scriptsize(\hphantom{-}+2.7)}
     & 57.47 & \textbf{66.87} \textcolor{LimeGreen}{\scriptsize(\hphantom{-}+9.4)}
     & \textbf{51.00} & 35.80 \textcolor{red}{\scriptsize(\hphantom{+}-15.2)} \\
2048 & 41.16 & \textbf{45.02} \textcolor{LimeGreen}{\scriptsize(\hphantom{-}+3.9)}
     & 6.49 & \textbf{8.93}  \textcolor{LimeGreen}{\scriptsize(\hphantom{-}+2.4)}
     & 62.62 & \textbf{69.83} \textcolor{LimeGreen}{\scriptsize(\hphantom{-}+7.2)}
     & \textbf{60.40} & 37.60 \textcolor{red}{\scriptsize(\hphantom{+}-22.8)} \\
4096 & \textbf{43.16} & \textbf{43.28} \textcolor{LimeGreen}{\scriptsize(\hphantom{-}+0.1)}
     & 8.33 & \textbf{10.22} \textcolor{LimeGreen}{\scriptsize(\hphantom{-}+1.9)}
     & 53.90 & \textbf{71.04} \textcolor{LimeGreen}{\scriptsize(\hphantom{-}+17.1)}
     & \textbf{50.80} & 41.00 \textcolor{red}{\scriptsize(\hphantom{+}-9.8)} \\
8192 & 39.98 & \textbf{43.72} \textcolor{LimeGreen}{\scriptsize(\hphantom{-}+3.7)}
     & \textbf{12.22} & \textbf{12.98} \textcolor{LimeGreen}{\scriptsize(\hphantom{-}+0.8)}
     & 41.70 & \textbf{70.89} \textcolor{LimeGreen}{\scriptsize(\hphantom{-}+29.2)}
     & 38.00 & \textbf{43.20} \textcolor{LimeGreen}{\scriptsize(\hphantom{-}+5.2)} \\
\bottomrule
\end{tabular}
}
\end{table*}

\begin{table*}[h]
\centering
\caption{
Baseline vs TRSD accuracy under truncated inference on out-of-distribution benchmarks (smaller/mid-size models).
Each block reports a fixed TRSD source dataset (shown in the block header) and evaluates on the remaining benchmarks.
Entries report Baseline/TRSD accuracy (higher is better). Colored values report the absolute change in accuracy
(TRSD $-$ Baseline, percentage points). Bold indicates the better entry; when Baseline and TRSD accuracies differ by at most
1 percentage point, both are bold.
}
\label{tab:small_models_out_combined}

\begin{tabular}{lcccccc}
\toprule

\multicolumn{7}{c}{\emph{Teacher Model: Qwen3-0.6B \;|\; TRSD: Countdown}} \\
\midrule
\multirow{2}{*}{Budget / Data} &
\multicolumn{2}{c}{Dolci} &
\multicolumn{2}{c}{GSM8K} &
\multicolumn{2}{c}{Math500} \\
\cmidrule(lr{4pt}){2-3}
\cmidrule(lr{4pt}){4-5}
\cmidrule(l{6pt}){6-7}
& Baseline & TRSD &
Baseline & TRSD &
Baseline & TRSD \\
\midrule
32   & \textbf{3.37} & \textbf{4.01} \textcolor{LimeGreen}{\scriptsize(\hphantom{-}+0.6)}
     & 3.03 & \textbf{5.23} \textcolor{LimeGreen}{\scriptsize(\hphantom{-}+2.2)}
     & 7.20 & \textbf{11.40} \textcolor{LimeGreen}{\scriptsize(\hphantom{-}+4.2)} \\
64   & \textbf{3.89} & \textbf{4.69} \textcolor{LimeGreen}{\scriptsize(\hphantom{-}+0.8)}
     & 3.41 & \textbf{5.46} \textcolor{LimeGreen}{\scriptsize(\hphantom{-}+2.1)}
     & 6.40 & \textbf{11.80} \textcolor{LimeGreen}{\scriptsize(\hphantom{-}+5.4)} \\
128  & 3.37 & \textbf{4.77} \textcolor{LimeGreen}{\scriptsize(\hphantom{-}+1.4)}
     & 4.47 & \textbf{12.36} \textcolor{LimeGreen}{\scriptsize(\hphantom{-}+7.9)}
     & 7.40 & \textbf{13.80} \textcolor{LimeGreen}{\scriptsize(\hphantom{-}+6.4)} \\
256  & 3.93 & \textbf{5.09} \textcolor{LimeGreen}{\scriptsize(\hphantom{-}+1.2)}
     & 23.96 & \textbf{41.17} \textcolor{LimeGreen}{\scriptsize(\hphantom{-}+17.2)}
     & 12.40 & \textbf{21.60} \textcolor{LimeGreen}{\scriptsize(\hphantom{-}+9.2)} \\
512  & \textbf{4.81} & \textbf{5.61} \textcolor{LimeGreen}{\scriptsize(\hphantom{-}+0.8)}
     & 59.14 & \textbf{64.22} \textcolor{LimeGreen}{\scriptsize(\hphantom{-}+5.1)}
     & 31.80 & \textbf{38.60} \textcolor{LimeGreen}{\scriptsize(\hphantom{-}+6.8)} \\
1024 & 5.05 & \textbf{7.05} \textcolor{LimeGreen}{\scriptsize(\hphantom{-}+2.0)}
     & \textbf{70.81} & \textbf{70.36} \textcolor{red}{\scriptsize(\hphantom{+}-0.5)}
     & 47.40 & \textbf{52.40} \textcolor{LimeGreen}{\scriptsize(\hphantom{-}+5.0)} \\
2048 & 6.85 & \textbf{8.09} \textcolor{LimeGreen}{\scriptsize(\hphantom{-}+1.2)}
     & \textbf{74.60} & 72.78 \textcolor{red}{\scriptsize(\hphantom{+}-1.8)}
     & \textbf{60.80} & 59.00 \textcolor{red}{\scriptsize(\hphantom{+}-1.8)} \\
4096 & \textbf{7.29} & \textbf{7.57} \textcolor{LimeGreen}{\scriptsize(\hphantom{-}+0.3)}
     & \textbf{76.35} & 73.16 \textcolor{red}{\scriptsize(\hphantom{+}-3.2)}
     & \textbf{67.00} & 63.40 \textcolor{red}{\scriptsize(\hphantom{+}-3.6)} \\
8192 & 6.77 & \textbf{7.97} \textcolor{LimeGreen}{\scriptsize(\hphantom{-}+1.2)}
     & \textbf{76.50} & 73.01 \textcolor{red}{\scriptsize(\hphantom{+}-3.5)}
     & \textbf{70.40} & 65.20 \textcolor{red}{\scriptsize(\hphantom{+}-5.2)} \\

\midrule
\multicolumn{7}{c}{\emph{Teacher Model: Qwen3-1.7B \;|\; TRSD: GSM8K}} \\
\midrule
\multirow{2}{*}{Budget / Data} &
\multicolumn{2}{c}{Countdown} &
\multicolumn{2}{c}{Dolci} &
\multicolumn{2}{c}{Math500} \\
\cmidrule(lr{4pt}){2-3}
\cmidrule(lr{4pt}){4-5}
\cmidrule(l{6pt}){6-7}
& Baseline & TRSD &
Baseline & TRSD &
Baseline & TRSD \\
\midrule
32   & \textbf{10.68} & 7.14 \textcolor{red}{\scriptsize(\hphantom{+}-3.5)}
     & 4.37 & \textbf{8.77} \textcolor{LimeGreen}{\scriptsize(\hphantom{-}+4.4)}
     & 11.60 & \textbf{15.80} \textcolor{LimeGreen}{\scriptsize(\hphantom{-}+4.2)} \\
64   & \textbf{13.50} & 9.64 \textcolor{red}{\scriptsize(\hphantom{+}-3.9)}
     & 3.89 & \textbf{8.93} \textcolor{LimeGreen}{\scriptsize(\hphantom{-}+5.0)}
     & 11.20 & \textbf{16.20} \textcolor{LimeGreen}{\scriptsize(\hphantom{-}+5.0)} \\
128  & \textbf{19.40} & 18.20 \textcolor{red}{\scriptsize(\hphantom{+}-1.2)}
     & 3.81 & \textbf{9.29} \textcolor{LimeGreen}{\scriptsize(\hphantom{-}+5.5)}
     & 12.20 & \textbf{17.80} \textcolor{LimeGreen}{\scriptsize(\hphantom{-}+5.6)} \\
256  & \textbf{39.24} & \textbf{39.64} \textcolor{LimeGreen}{\scriptsize(\hphantom{-}+0.4)}
     & 4.29 & \textbf{9.78} \textcolor{LimeGreen}{\scriptsize(\hphantom{-}+5.5)}
     & 17.80 & \textbf{25.80} \textcolor{LimeGreen}{\scriptsize(\hphantom{-}+8.0)} \\
512  & \textbf{55.60} & \textbf{55.14} \textcolor{red}{\scriptsize(\hphantom{+}-0.5)}
     & 6.57 & \textbf{10.98} \textcolor{LimeGreen}{\scriptsize(\hphantom{-}+4.4)}
     & 38.00 & \textbf{46.00} \textcolor{LimeGreen}{\scriptsize(\hphantom{-}+8.0)} \\
1024 & \textbf{66.10} & \textbf{65.12} \textcolor{red}{\scriptsize(\hphantom{+}-1.0)}
     & 8.17 & \textbf{12.94} \textcolor{LimeGreen}{\scriptsize(\hphantom{-}+4.8)}
     & 60.00 & \textbf{64.20} \textcolor{LimeGreen}{\scriptsize(\hphantom{-}+4.2)} \\
2048 & \textbf{72.60} & 70.20 \textcolor{red}{\scriptsize(\hphantom{+}-2.4)}
     & 9.82 & \textbf{15.30} \textcolor{LimeGreen}{\scriptsize(\hphantom{-}+5.5)}
     & \textbf{74.20} & 72.60 \textcolor{red}{\scriptsize(\hphantom{+}-1.6)} \\
4096 & \textbf{77.54} & 73.90 \textcolor{red}{\scriptsize(\hphantom{+}-3.6)}
     & 12.78 & \textbf{19.91} \textcolor{LimeGreen}{\scriptsize(\hphantom{-}+7.1)}
     & \textbf{80.40} & \textbf{79.80} \textcolor{red}{\scriptsize(\hphantom{+}-0.6)} \\
8192 & \textbf{80.16} & 74.70 \textcolor{red}{\scriptsize(\hphantom{+}-5.5)}
     & 17.23 & \textbf{25.40} \textcolor{LimeGreen}{\scriptsize(\hphantom{-}+8.2)}
     & \textbf{84.60} & 81.20 \textcolor{red}{\scriptsize(\hphantom{+}-3.4)} \\

\midrule
\multicolumn{7}{c}{\emph{Teacher Model: OpenThinker3-1.5B \;|\; TRSD: GSM8K}} \\
\midrule
\multirow{2}{*}{Budget / Data} &
\multicolumn{2}{c}{Countdown} &
\multicolumn{2}{c}{Dolci} &
\multicolumn{2}{c}{Math500} \\
\cmidrule(lr{4pt}){2-3}
\cmidrule(lr{4pt}){4-5}
\cmidrule(l{6pt}){6-7}
& Baseline & TRSD &
Baseline & TRSD &
Baseline & TRSD \\
\midrule
32   & 0.26 & \textbf{2.34} \textcolor{LimeGreen}{\scriptsize(\hphantom{-}+2.1)}
     & 3.73 & \textbf{4.93} \textcolor{LimeGreen}{\scriptsize(\hphantom{-}+1.2)}
     & 3.40 & \textbf{10.20} \textcolor{LimeGreen}{\scriptsize(\hphantom{-}+6.8)} \\
64   & 0.46 & \textbf{2.90} \textcolor{LimeGreen}{\scriptsize(\hphantom{-}+2.4)}
     & 4.49 & \textbf{5.97} \textcolor{LimeGreen}{\scriptsize(\hphantom{-}+1.5)}
     & 6.00 & \textbf{10.00} \textcolor{LimeGreen}{\scriptsize(\hphantom{-}+4.0)} \\
128  & 0.82 & \textbf{2.44} \textcolor{LimeGreen}{\scriptsize(\hphantom{-}+1.6)}
     & 4.37 & \textbf{6.33} \textcolor{LimeGreen}{\scriptsize(\hphantom{-}+2.0)}
     & 6.20 & \textbf{12.40} \textcolor{LimeGreen}{\scriptsize(\hphantom{-}+6.2)} \\
256  & 4.84 & \textbf{7.78} \textcolor{LimeGreen}{\scriptsize(\hphantom{-}+2.9)}
     & 3.37 & \textbf{6.25} \textcolor{LimeGreen}{\scriptsize(\hphantom{-}+2.9)}
     & 11.20 & \textbf{15.80} \textcolor{LimeGreen}{\scriptsize(\hphantom{-}+4.6)} \\
512  & 16.64 & \textbf{18.86} \textcolor{LimeGreen}{\scriptsize(\hphantom{-}+2.2)}
     & 4.65 & \textbf{7.25} \textcolor{LimeGreen}{\scriptsize(\hphantom{-}+2.6)}
     & 24.80 & \textbf{32.80} \textcolor{LimeGreen}{\scriptsize(\hphantom{-}+8.0)} \\
1024 & 28.28 & \textbf{31.02} \textcolor{LimeGreen}{\scriptsize(\hphantom{-}+2.7)}
     & 4.53 & \textbf{9.33} \textcolor{LimeGreen}{\scriptsize(\hphantom{-}+4.8)}
     & \textbf{51.00} & \textbf{50.20} \textcolor{red}{\scriptsize(\hphantom{+}-0.8)} \\
2048 & \textbf{41.16} & 32.64 \textcolor{red}{\scriptsize(\hphantom{+}-8.5)}
     & 6.49 & \textbf{10.78} \textcolor{LimeGreen}{\scriptsize(\hphantom{-}+4.3)}
     & \textbf{60.40} & 55.20 \textcolor{red}{\scriptsize(\hphantom{+}-5.2)} \\
4096 & \textbf{43.16} & 32.00 \textcolor{red}{\scriptsize(\hphantom{+}-11.2)}
     & 8.33 & \textbf{13.10} \textcolor{LimeGreen}{\scriptsize(\hphantom{-}+4.8)}
     & 50.80 & \textbf{59.20} \textcolor{LimeGreen}{\scriptsize(\hphantom{-}+8.4)} \\
8192 & \textbf{39.98} & 33.34 \textcolor{red}{\scriptsize(\hphantom{+}-6.6)}
     & 12.22 & \textbf{16.59} \textcolor{LimeGreen}{\scriptsize(\hphantom{-}+4.4)}
     & 38.00 & \textbf{59.80} \textcolor{LimeGreen}{\scriptsize(\hphantom{-}+21.8)} \\
\bottomrule
\end{tabular}
\end{table*}

\clearpage
\subsection{Reasoning length analysis for smaller models}
\label{appD:thinklen_small}

We complement the main analysis in Section~\ref{sec:emergent_shorter_reasoning} by examining how TRSD affects inference-time reasoning length for smaller model architectures.
Table \ref{tab:appendix_thinklen_small_models_combined} reports the average number of reasoning tokens generated under unconstrained inference ($r_{\max}=8192$), conditioned on answer correctness, for Qwen3-0.6B, Qwen3-1.7B, and OpenThinker3-1.5B.

We report in the top part results using per-dataset in-distribution checkpoints, following the same selection procedure as in the main experiments, while in the bottom part we fix a single TRSD checkpoint per model across all datasets, mirroring the out-of-distribution evaluation protocol.
Across all three models and datasets, the same qualitative pattern observed for larger architectures persists, where TRSD reduces reasoning length for both correct and incorrect predictions across nearly all settings. 
\begin{table}[t]
\centering
\caption{
Average number of reasoning tokens conditioned on answer correctness for smaller models.
Top block: in-distribution TRSD using per-dataset checkpoints selected as in Table~\ref{app:ckpt}.
Bottom block: out-of-distribution-style TRSD using a single fixed checkpoint per model
(Countdown step~500 for Qwen3-0.6B, GSM8K step~250 for Qwen3-1.7B, GSM8K step~2000 for OpenThinker3-1.5B).
Bold indicates the configuration (Baseline or TRSD) with lower average reasoning length.
}
\label{tab:appendix_thinklen_small_models_combined}
\resizebox{\columnwidth}{!}{
\begin{tabular}{@{}llcccccccc@{}}
\toprule
& & \multicolumn{2}{c}{Countdown} & \multicolumn{2}{c}{Dolci} & \multicolumn{2}{c}{GSM8K} & \multicolumn{2}{c}{Math500} \\
\cmidrule(lr){3-4}\cmidrule(lr){5-6}\cmidrule(lr){7-8}\cmidrule(lr){9-10}
Model & Outcome
& Baseline & TRSD
& Baseline & TRSD
& Baseline & TRSD
& Baseline & TRSD \\
\midrule

\multicolumn{10}{c}{\textit{In-distribution evaluation}} \\
\addlinespace[0.2em]
\midrule

\multirow{2}{*}{Qwen3-0.6B}
& Correct
& 1169 & \textbf{724}
& 5655 & \textbf{3765}
& 848  & \textbf{575}
& 2460 & \textbf{1575} \\
& Wrong
& 4719 & \textbf{3230}
& 7295 & \textbf{5268}
& 2183 & \textbf{1888}
& 5606 & \textbf{4506} \\
\addlinespace[0.15em]
\midrule
\addlinespace[0.15em]

\multirow{2}{*}{Qwen3-1.7B}
& Correct
& 1098 & \textbf{747}
& 6633 & \textbf{4590}
& 1005 & \textbf{710}
& 2879 & \textbf{2227} \\
& Wrong
& 6379 & \textbf{2482}
& 7777 & \textbf{6071}
& 3364 & \textbf{2501}
& 5829 & \textbf{5633} \\
\addlinespace[0.15em]
\midrule
\addlinespace[0.15em]

\multirow{2}{*}{OpenThinker3-1.5B}
& Correct
& 4814 & \textbf{2641}
& 7765 & \textbf{7375}
& 3679 & \textbf{1543}
& 4142 & \textbf{3625} \\
& Wrong
& 6517 & \textbf{5951}
& 7999 & \textbf{7788}
& 4720 & \textbf{3143}
& 4692 & \textbf{3571} \\

\addlinespace[0.45em]
\cmidrule(lr){1-10}
\addlinespace[0.25em]

\multicolumn{10}{c}{\textit{Out-of-distribution evaluation (fixed TRSD checkpoint)}} \\
\addlinespace[0.2em]
\midrule

\multirow{2}{*}{Qwen3-0.6B}
& Correct
& 1169 & \textbf{725}
& 5655 & \textbf{3833}
& 848  & \textbf{555}
& 2460 & \textbf{1738} \\
& Wrong
& 4719 & \textbf{3230}
& 7295 & \textbf{5750}
& 2183 & \textbf{1308}
& 5606 & \textbf{4371} \\
\addlinespace[0.15em]
\midrule
\addlinespace[0.15em]

\multirow{2}{*}{Qwen3-1.7B}
& Correct
& 1098 & \textbf{915}
& 6633 & \textbf{6362}
& 1005 & \textbf{710}
& 2879 & \textbf{2283} \\
& Wrong
& 6379 & \textbf{4518}
& 7777 & \textbf{7512}
& 3364 & \textbf{2501}
& 5829 & \textbf{5055} \\
\addlinespace[0.15em]
\midrule
\addlinespace[0.15em]

\multirow{2}{*}{OpenThinker3-1.5B}
& Correct
& 4814 & \textbf{2580}
& 7765 & \textbf{6806}
& 3679 & \textbf{1543}
& 4142 & \textbf{2876} \\
& Wrong
& 6517 & \textbf{5017}
& 7999 & \textbf{7642}
& 4720 & \textbf{3143}
& 4692 & \textbf{3951} \\
\bottomrule
\end{tabular}
}
\end{table}
\clearpage
\subsection{Per-dataset performance plots}
\label{appD:plots}

\begin{figure}[h]
    \centering
    \subfloat[Countdown]{\includegraphics[width=0.48\linewidth]{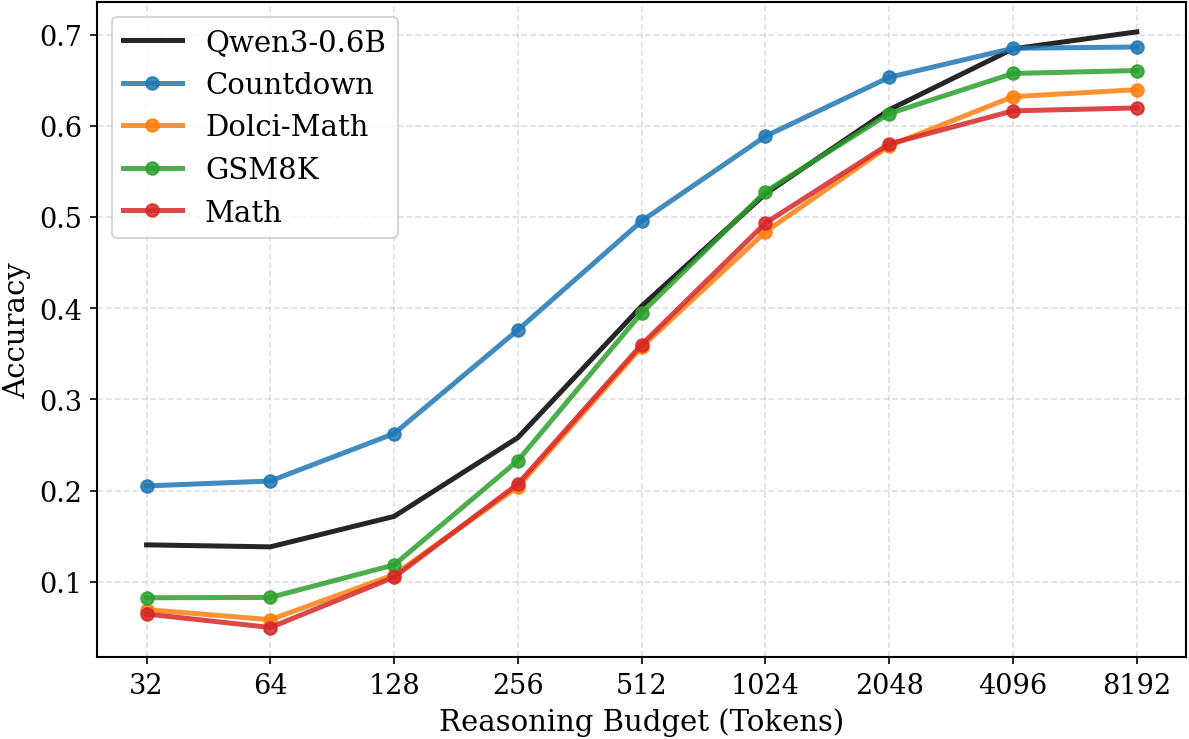}}
    \hfill
    \subfloat[Dolci]{\includegraphics[width=0.48\linewidth]{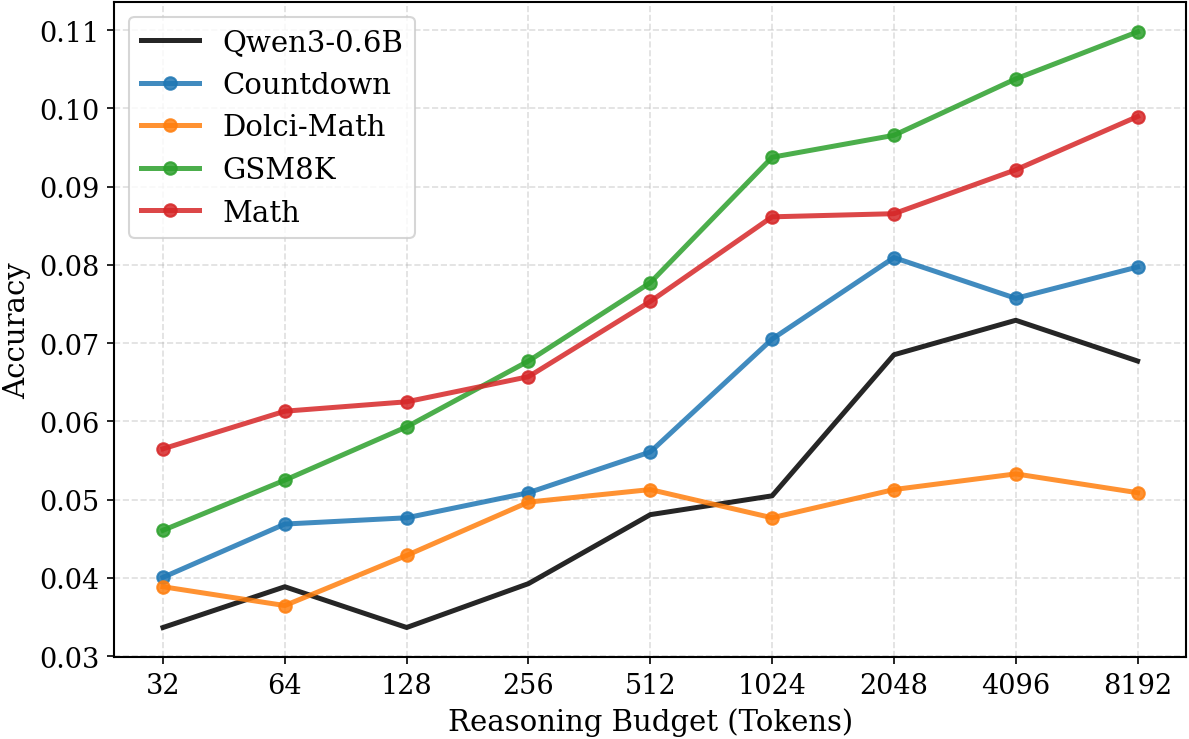}} \\
    \subfloat[GSM8K]{\includegraphics[width=0.48\linewidth]{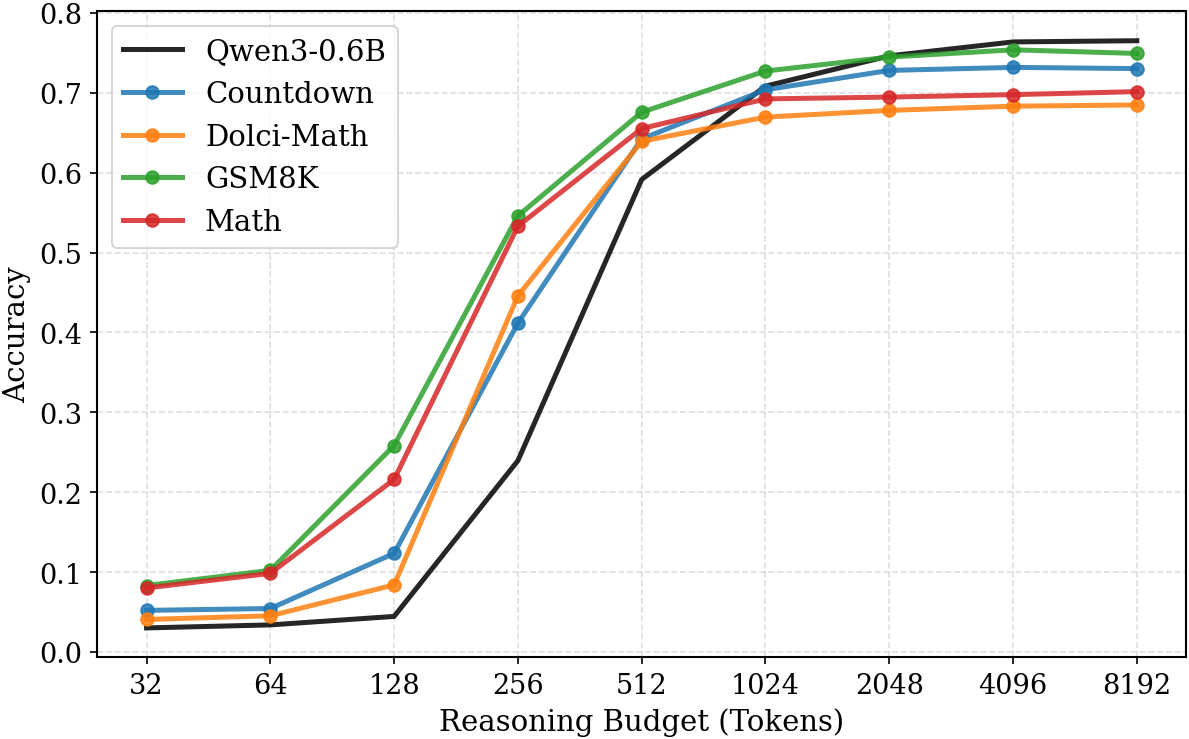}}
    \hfill
    \subfloat[Math500]{\includegraphics[width=0.48\linewidth]{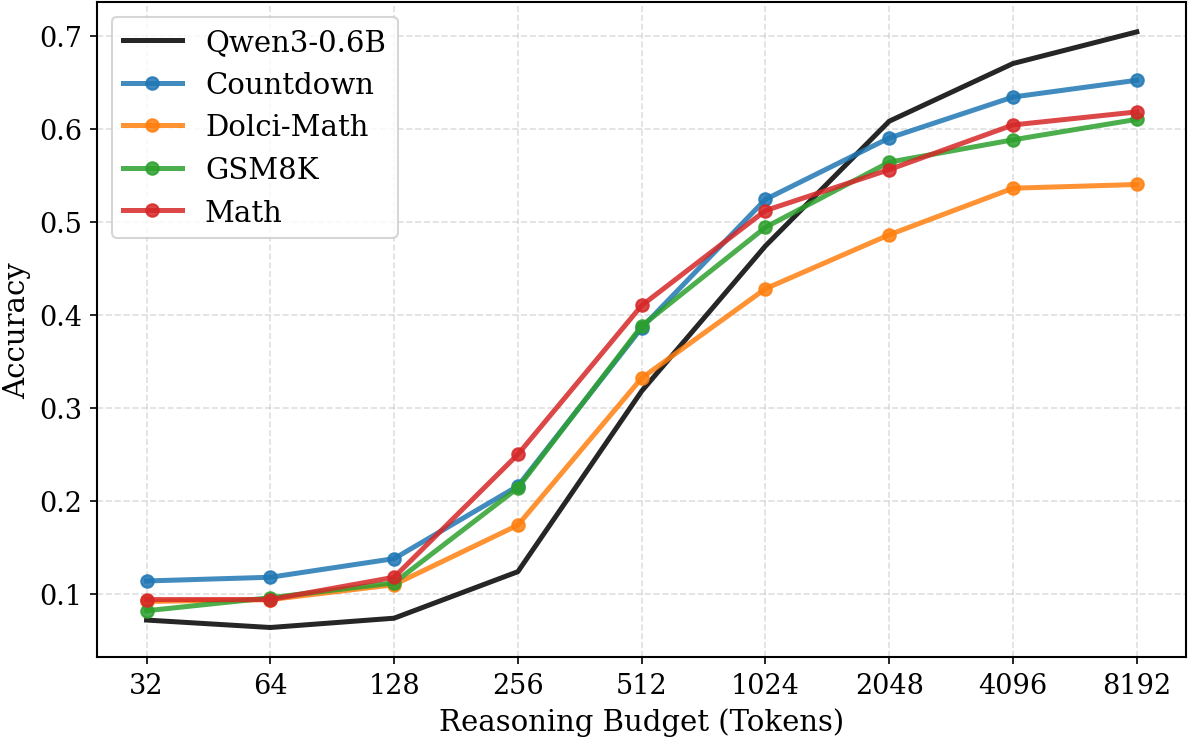}}
    \caption{Per-dataset accuracy as a function of the reasoning budget for Qwen3-0.6B. The evaluation dataset is specified below the respective plot.}
    \label{fig:perdata_qwen3_06b}
\end{figure}

\begin{figure}[h]
    \centering
    \subfloat[Countdown]{\includegraphics[width=0.48\linewidth]{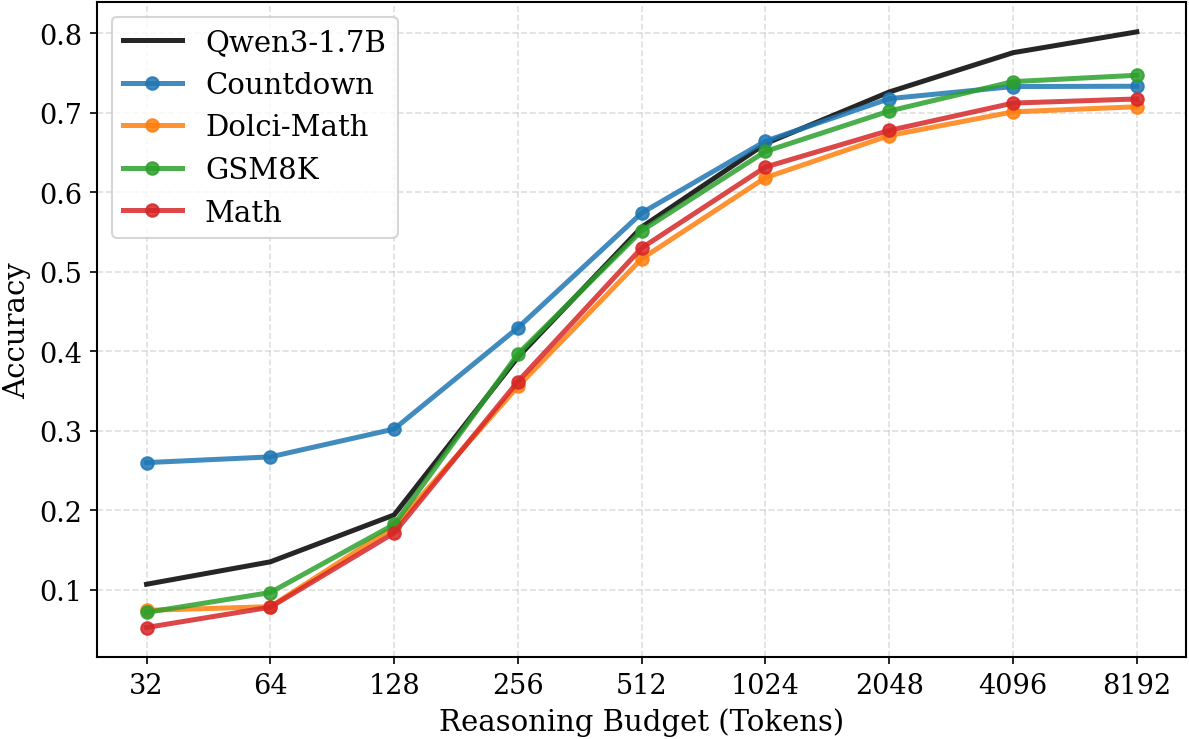}}
    \hfill
    \subfloat[Dolci]{\includegraphics[width=0.48\linewidth]{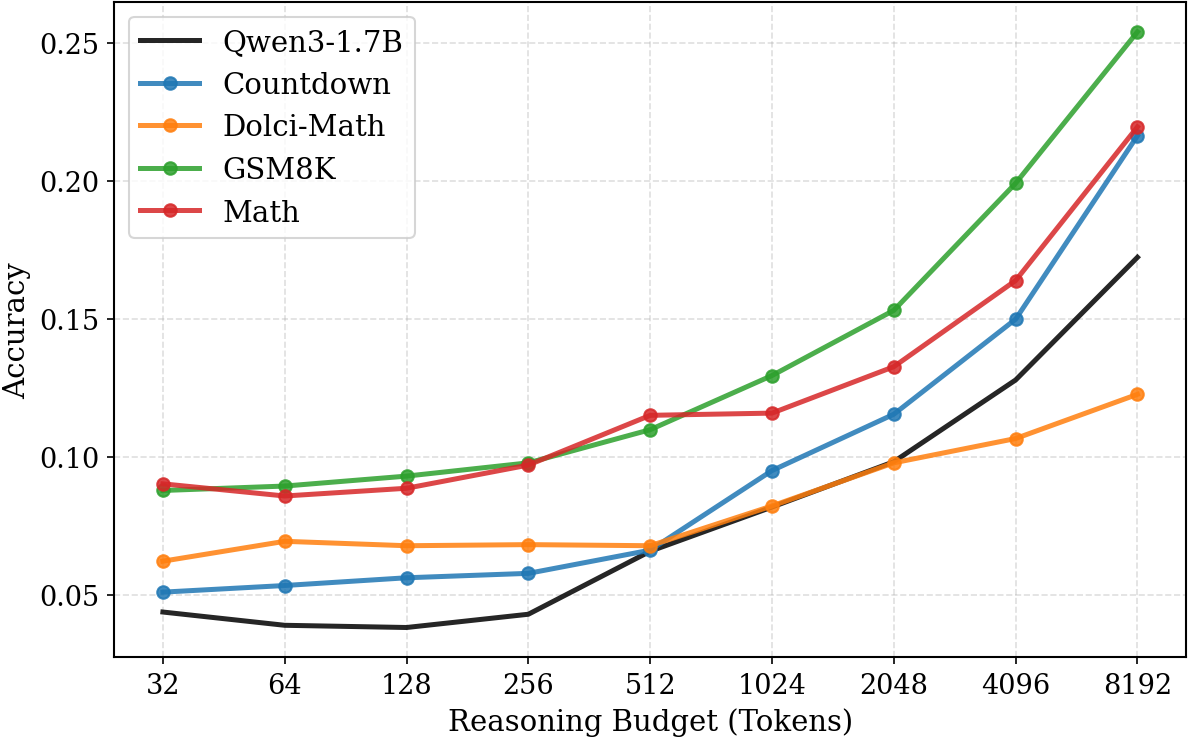}} \\
    \subfloat[GSM8K]{\includegraphics[width=0.48\linewidth]{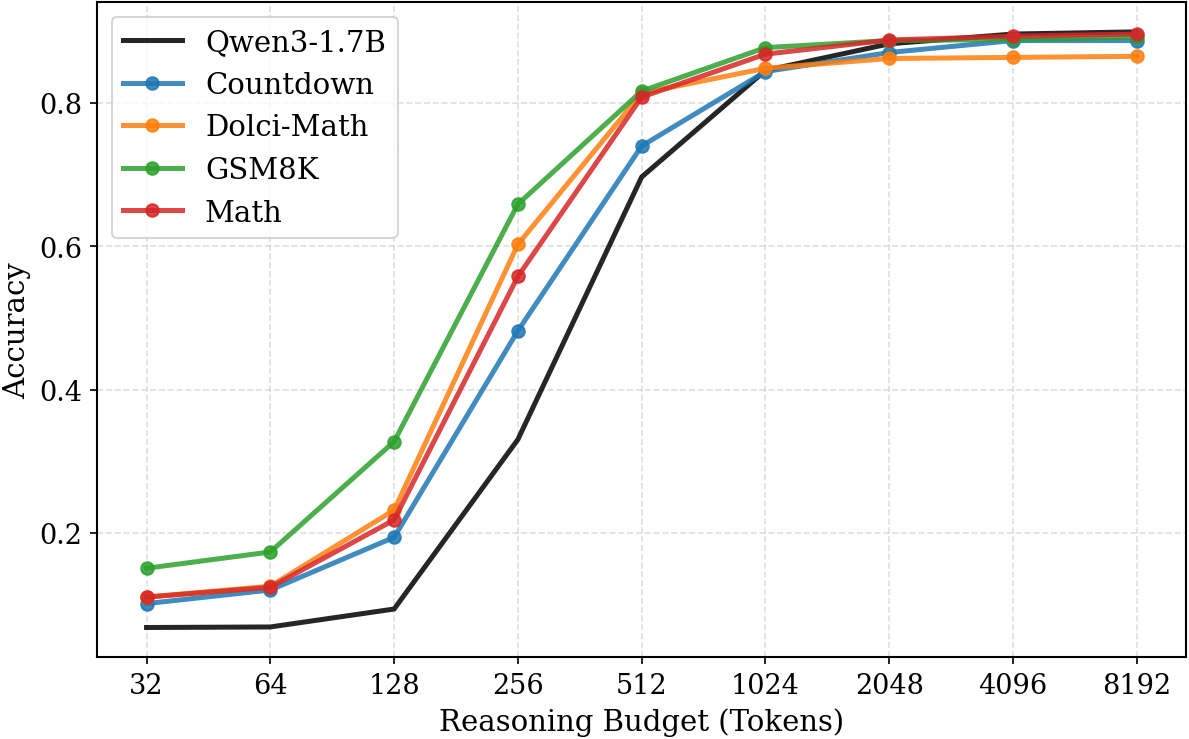}}
    \hfill
    \subfloat[Math500]{\includegraphics[width=0.48\linewidth]{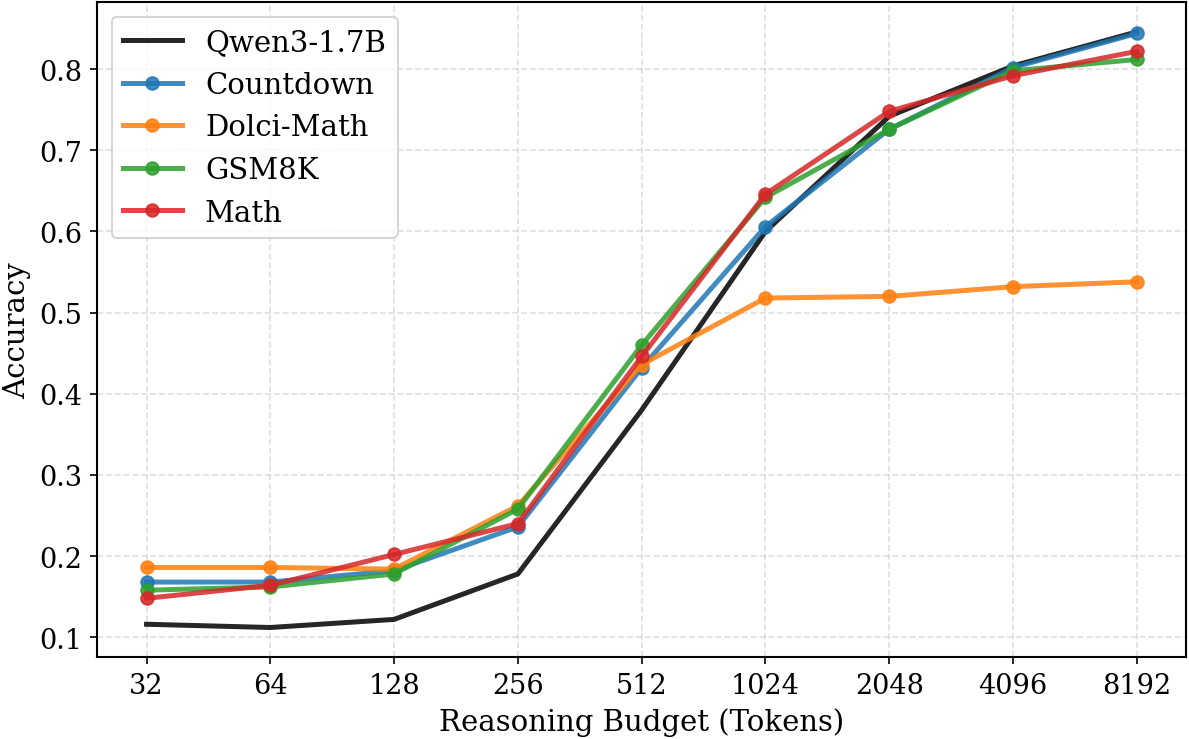}}
    \caption{Per-dataset accuracy as a function of the reasoning budget for Qwen3-1.7B. The evaluation dataset is specified below the respective plot.}
    \label{fig:perdata_qwen3_17b}
\end{figure}

\begin{figure}[h]
    \centering
    \subfloat[Countdown]{\includegraphics[width=0.48\linewidth]{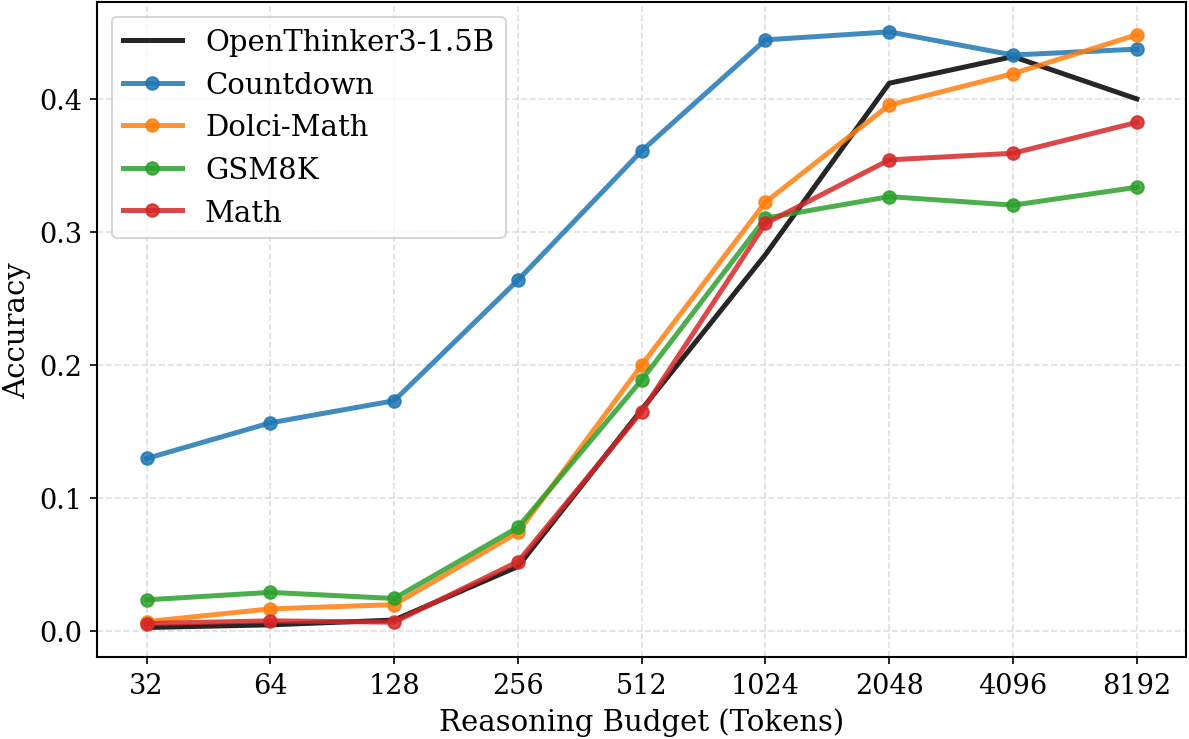}}
    \hfill
    \subfloat[Dolci]{\includegraphics[width=0.48\linewidth]{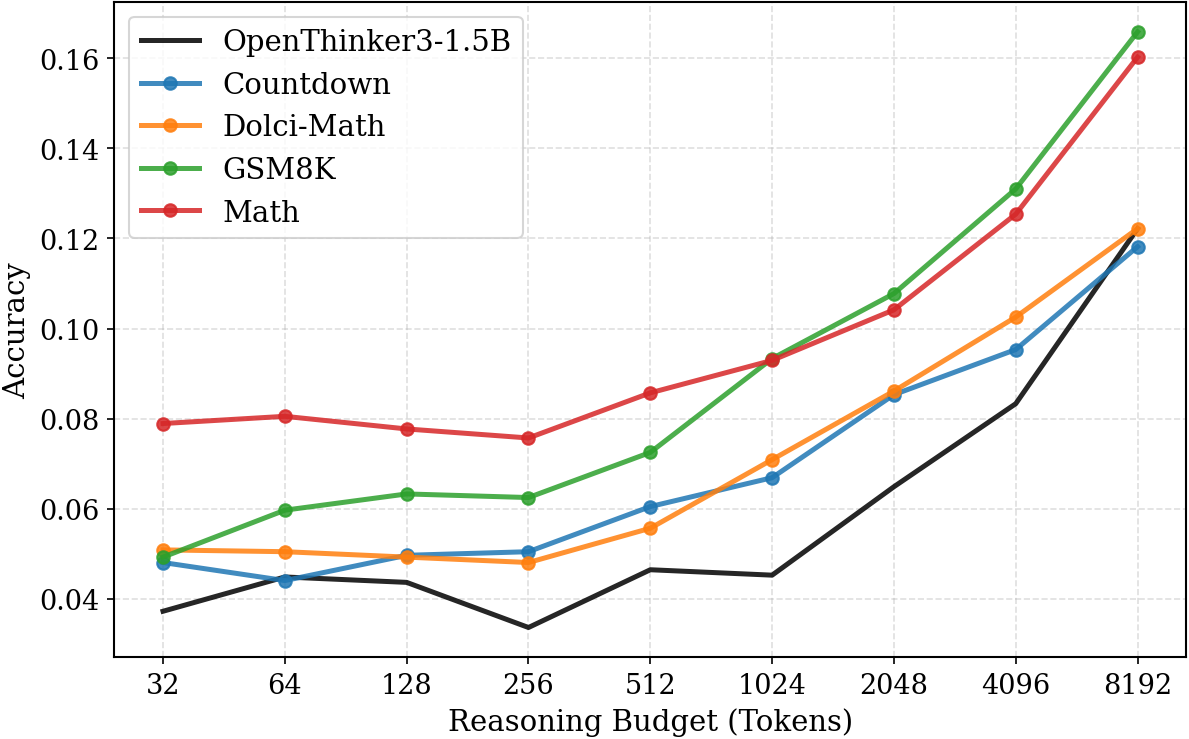}} \\
    \subfloat[GSM8K]{\includegraphics[width=0.48\linewidth]{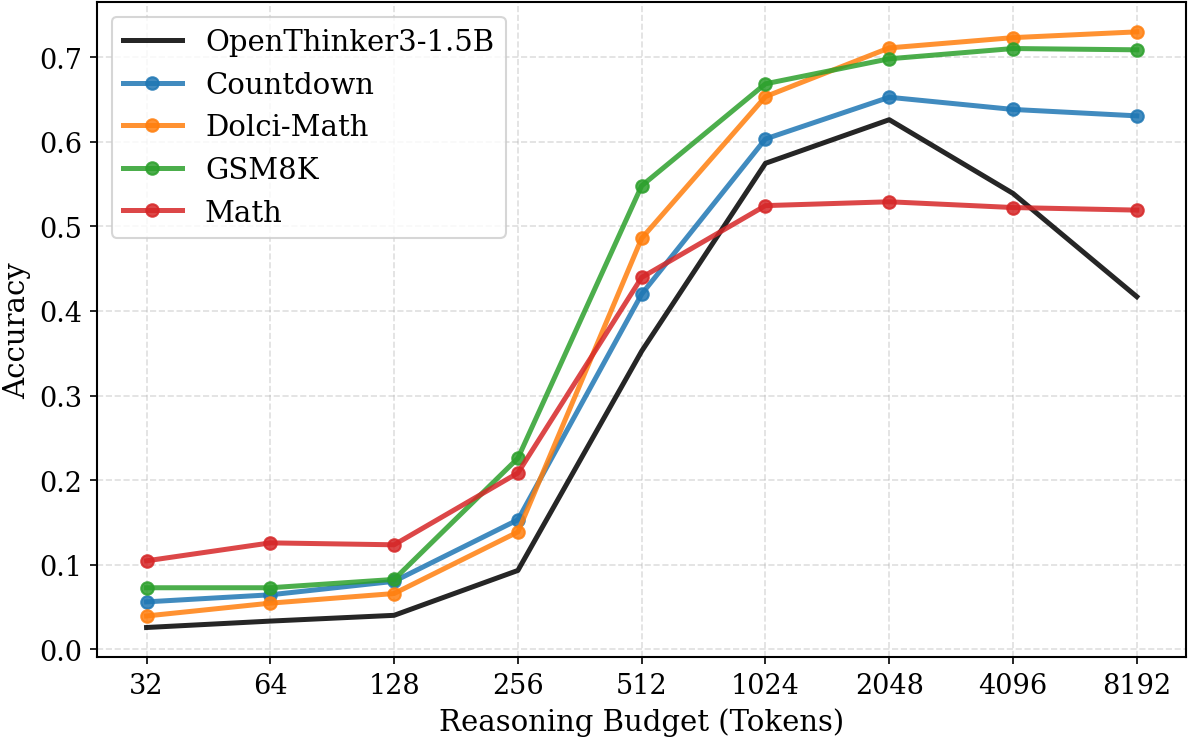}}
    \hfill
    \subfloat[Math500]{\includegraphics[width=0.48\linewidth]{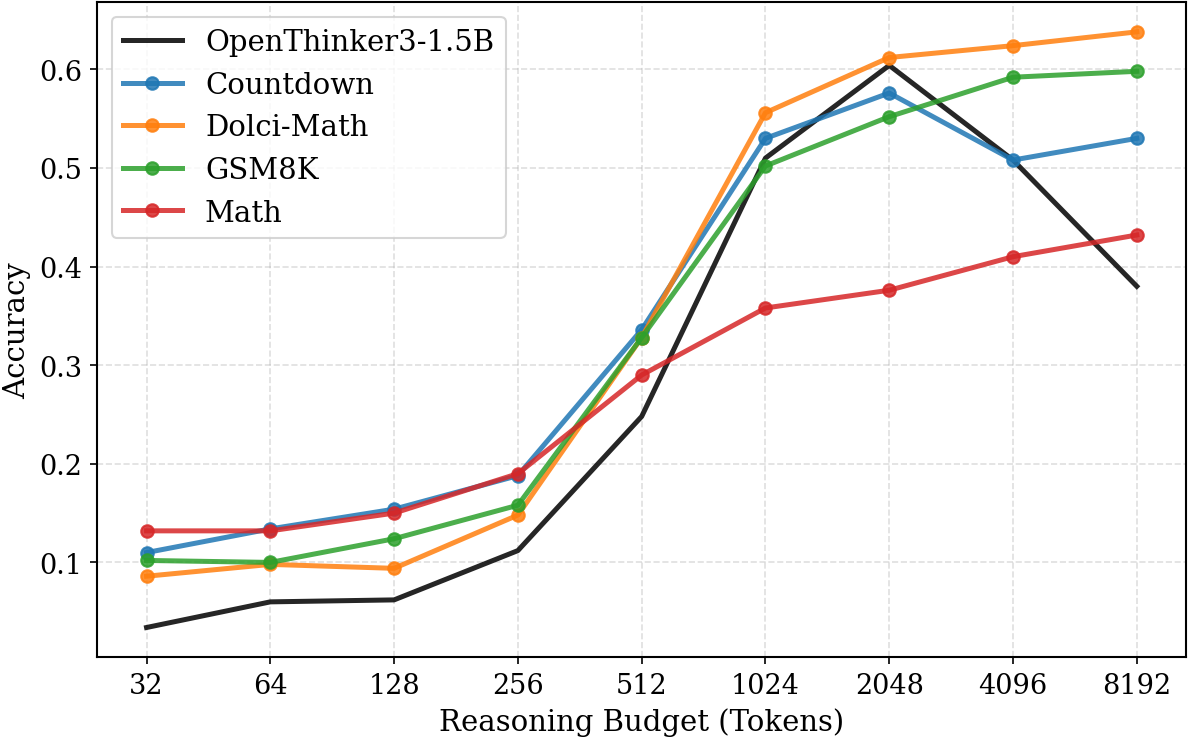}}
    \caption{Per-dataset accuracy as a function of the reasoning budget for OpenThinker3-1.5B. The evaluation dataset is specified below the respective plot.}
    \label{fig:perdata_openthinker3}
\end{figure}

To complement the aggregate tables, Figures~\ref{fig:perdata_qwen3_06b}, \ref{fig:perdata_qwen3_17b}, \ref{fig:perdata_phi4mini} and \ref{fig:perdata_openthinker3} visualize per-dataset performance as a function of the available reasoning budget. These plots make two consistent behaviors particularly clear.
First, the advantage of TRSD is concentrated in the low-budget regime. Across datasets, TRSD curves typically rise earlier or degrade more slowly as the budget decreases, indicating that distilled models can recover correct predictions from much shorter prefixes of the reasoning trace. In contrast, baseline models often exhibit sharp transitions in accuracy as the reasoning budget increases. Second, the plots highlight that TRSD does not uniformly improve performance across all budgets.
At moderate and large budgets, TRSD-trained models generally approach the baseline, and in some cases plateau below it at the largest budget.

\clearpage
\section{Additional Qualitative Examples}
\label{app:qualitative}

We provide additional qualitative comparisons between baseline and TRSD-trained models.
We consider two complementary regimes: cases where \emph{both models answer correctly}, and cases
where \emph{both models fail on the same prompt}. These examples are intended to illustrate
differences in reasoning behavior and representative shared failure modes, rather than to suggest
systematic correctness improvements beyond those reported in the main paper. In both cases, we use Qwen3-4B as baseline and the TRSD version trained with prompts from GSM8K.

\paragraph{Both models correct.}
Figure \ref{fig:qualitative_correct_appendix} shows a representative example where both the baseline
and TRSD-trained models produce the correct answer on the Countdown dataset. While correctness is preserved, the TRSD-trained
model typically reaches the solution using a shorter and more focused reasoning trace, whereas the
baseline exhibits additional intermediate steps or redundant self-verification. This behavior
mirrors the quantitative trends reported in Section \ref{sec4:experiments}, where TRSD reduces
reasoning length without degrading accuracy.

\paragraph{Both models incorrect.}
Figure \ref{fig:qualitative_wrong_1} presents a case where both
models fail on the same prompt from GSM8K. These examples illustrate shared failure modes, such as semantic
ambiguity or invalid implicit assumptions, that are not directly addressed by reasoning truncation.
\begin{figure}[t]
\centering
\tiny
\setlength{\tabcolsep}{6pt}
\scalebox{0.9}{
\begin{tcolorbox}[
    colback=gray!3,
    colframe=gray!50,
    boxrule=0.5pt,
    arc=2pt,
    title={Qualitative comparison of reasoning traces (Qwen3-4B, GSM8K)}
]

\textbf{Prompt.}
Using the numbers 59, 92, 24, 62, create an equation that equals 53. You can use basic arithmetic operations (+, -, *, /) one or multiple times but each number can only be used once.

\vspace{0.6em}

\begin{tabular}{p{0.46\linewidth} p{0.46\linewidth}}
\textbf{Baseline (Qwen3-4B)} &
\textbf{TRSD (GSM8K)} \\[0.4em]

\begin{minipage}[t]{\linewidth}
\textbf{Think:}\\
Okay, let's see. I need to use the numbers 59, 92, 24, and 62 exactly once each, and combine them with basic arithmetic operations to make 53. Hmm. Let me start by thinking about possible operations.

First, maybe I can subtract some numbers. Let's see, 92 is a big number. If I subtract something from 92, maybe that could help. Let me try 92 minus something. For example, 92 minus 59 is 33. But then I have 24 and 62 left. Not sure how to combine those. Maybe 24 plus 62 is 86. 33 and 86... that's way over. Not helpful.

Alternatively, maybe 62 minus something. Let's see. If I take 62 and subtract 24, that's 38. Then maybe 92 minus 59 is 33. 38 and 33... not helpful. Hmm.

Wait, maybe I need to use division or multiplication. Let me think. For example, 59 minus something. If I can get 59 minus a number that equals 53, but that would require subtracting 6, but I don't have a 6. Alternatively, maybe combining numbers through operations.

What if I do 92 minus 62? That's 30. Then maybe 59 minus 24 is 35. 30 and 35... not helpful. Or 30 plus 35 is 65. Not 53.

Another approach: Maybe multiply two numbers and then subtract others. Let's see. 24 times something. 24 times 2 is 48, but I don't have a 2. Alternatively, 24 times 59? That's way too big. 24*59 is 1416. Not helpful.

How about division? Let's see. Maybe 92 divided by something. 92 divided by 2 is 46, but again, no 2. Wait, 62 divided by something. Maybe 62 divided by 24? That's around 2.58. Not sure.

Alternatively, maybe 59 minus (something). Let's think. If I can get 59 minus (something) equals 53, that something would be 6. But how to get 6 from 92, 24, and 62? Let me try. 92 minus 62 is 30. Then 30 minus 24 is 6. Oh! Wait, that's 92 - 62 = 30, then 30 - 24 = 6. So then 59 - 6 = 53. But wait, that uses 92, 62, 24, and 59. Each number is used once. Let me check:

So 59 - ((92 - 62) - 24) = 59 - (30 - 24) = 59 - 6 = 53. Yes! That works. Let me verify the operations step by step.

First, compute 92 - 62 = 30. Then subtract 24 from that: 30 - 24 = 6. Then subtract that result from 59: 59 - 6 = 53. Yes! So that uses all four numbers once each. Let me write that as an equation.

So the equation would be 59 - ((92 - 62) - 24) = 53. Alternatively, simplifying the parentheses, it's 59 - 92 + 62 - 24. Wait, let me check that. Because (92 - 62) is 30, then subtract 24 gives 6. So 59 - 6 is 53. Alternatively, expanding the parentheses: 59 - 92 + 62 - 24. Let me compute that: 59 -92 is -33, plus 62 is 29, minus 24 is 5. Wait, that's not 53. Wait, that can't be right. Wait, what's wrong here?

Wait, no. The original expression is 59 - ((92 - 62) - 24). Which is 59 - ( (92 - 62) - 24 ). Let's compute the inner part first. 92 - 62 is 30. Then subtract 24: 30 -24 is 6. Then 59 - 6 is 53. So that's correct. But if I expand the parentheses, it's 59 - 92 + 62 - 24? Wait, no. Because ( (92 - 62) -24 ) is 92 -62 -24. So the entire expression is 59 - (92 -62 -24) = 59 -92 +62 +24? Wait, no. Wait, subtracting a sum is the same as adding the negatives. So 59 - (92 -62 -24) = 59 -92 +62 +24. Let me compute that: 59 -92 is -33, plus 62 is 29, plus 24 is 53. Oh right! So that's correct. So the equation can be written as 59 - 92 + 62 + 24. But wait, that would be 59 -92 is -33, plus 62 is 29, plus 24 is 53. Yes. So that's another way to write it. But in that case, the original expression is 59 - ( (92 -62) -24 ). Which simplifies to 59 -92 +62 +24. But that uses all four numbers once. So that's valid.

Alternatively, maybe the user wants the equation to be written with parentheses. So the answer would be 59 - ((92 - 62) - 24) = 53. But I need to check if that's allowed. The problem says each number can be used once. So yes, 59, 92, 62, 24 are all used once. So that works.

Alternatively, maybe there's another way. Let me check. For example, 62 - 24 = 38. 92 - 59 = 33. 38 + 33 = 71. Not helpful. Or 92 - 59 = 33. 62 - 24 = 38. 33 + 38 = 71. Still no. How about 92 - (62 + 24) = 92 - 86 = 6. Then 59 - 6 = 53. Oh! That's another way. So that would be 59 - (92 - (62 + 24)) = 53. Let's check: 62 +24 is 86. 92 -86 is 6. 59 -6 is 53. So that's another valid equation. Which one is better? Both are correct. Either way.

So the equation can be written as 59 - (92 - (62 + 24)) = 53. Or 59 - ((92 - 62) - 24) = 53. Either is acceptable.

Alternatively, maybe using multiplication or division. Let me think. For example, 24 divided by something. 24 / (something) to get a number. But I don't see a straightforward way. Let me try. Suppose I do 92 - (59 + 24) = 92 - 83 = 9. Then 62 - 9 = 53. But that would be 62 - (92 - (59 +24)) = 53. Let me check: 59 +24 is 83. 92 -83 is 9. 62 -9 is 53. So that's another way. But that uses all numbers once. So 62 - (92 - (59 +24)) = 53. That's another valid equation.

So there are multiple correct answers. The problem just requires one. So any of these would work. Let me pick one. The first one I thought of was 59 - ((92 -62) -24) =53. Alternatively, the second one 59 - (92 - (62 +24)) =53. Either is correct.

But let me check if the numbers are all used once. In the first case: 59, 92, 62, 24. Yes. In the second case, same numbers. So either is correct. The answer can be written in any of these forms. So perhaps the simplest way is to write 59 - (92 - 62 - 24) = 53. Wait, but that would be 59 - (92 -62 -24). Let's compute that. 92 -62 is 30, minus 24 is 6. 59 -6 is 53. So that's the same as before.

Alternatively, 59 - 92 + 62 +24. Which is also correct. But the problem says to use the numbers once each. So that's acceptable.

But maybe the problem expects parentheses to show the order of operations. So perhaps the answer is written as 59 - (92 - 62 - 24) = 53. Or maybe 59 - ( (92 - 62) - 24 ) = 53. Either is fine.

I think that's the correct approach. So the final answer is 59 - (92 - 62 - 24) = 53. Let me check again:

Compute 92 -62 =30. Then 30 -24 =6. Then 59 -6=53. Correct.

Alternatively, 92 -62 -24 = 6. 59 -6=53. Yes. So that's valid. So the equation is 59 - (92 -62 -24) =53. Or written with parentheses as needed.

So the answer would be that equation. Let me make sure I didn't miss any other possible ways. Another way could be 62 - (92 - (59 +24)) =53. Let me check: 59+24=83. 92-83=9. 62-9=53. Yes. So that's another valid equation. But the problem didn't specify any preference, so any of these is acceptable.

But the first one I found is probably easier. So I think the answer is 59 - (92 - 62 - 24) = 53.

\vspace{0.4em}
\textbf{Answer:}
59 - (92 - 62 - 24)
\end{minipage}
&
\begin{minipage}[t]{\linewidth}
\textbf{Think:}\\
Okay, let's see. I need to use the numbers 59, 92, 24, and 62 exactly once each with basic arithmetic operations to make 53. Hmm. Let me start by thinking about possible operations.

First, maybe subtract some numbers. Let's see, 92 minus 59 is 33. Then 62 minus 24 is 38. Not sure how that helps. Wait, maybe combining them differently. Let me try adding some numbers and subtracting others.

What if I take 92 and subtract 62? That gives 30. Then maybe add 24? 30 +24 is 54. Close to 53, but one off. Oh, but I still have 59 left. Wait, but I can't use numbers more than once. Wait, I already used 92, 62, and 24. So 59 is left. Maybe that's not the right path.

Alternatively, let's try 59 minus something. Let's see, 59 - (92 - 62) = 59 -30 =29. Then adding 24? 29 +24=53. Wait, but that would use 59, 92, 62, and 24. Each exactly once. Let me check that.

So 92 -62 is 30. Then 59 -30 is 29. Then 29 +24 is 53. Wait, but how to structure that into an equation using all numbers once. Let me write it step by step:

First, 92 - 62 = 30

Then 59 - 30 = 29

Then 29 +24 =53.

But how to combine these steps into one equation. Let's see:

(59 - (92 - 62)) +24 =53.

Yes, that uses all numbers once. Let me verify:

Compute inside the parentheses first: 92 -62 is 30. Then 59 -30 is 29. Then add 24: 29 +24 =53. Perfect. So the equation would be (59 - (92 - 62)) +24 =53.

Alternatively, maybe there's another way, but this seems to work. Let me check if each number is used once: 59, 92, 62, 24. Yes, all four numbers are used once each. The operations are subtraction and addition. So that's valid.

\vspace{0.4em}
\textbf{Answer:}
 (59 - (92 - 62)) + 24
\end{minipage}
\end{tabular}

\end{tcolorbox}
}

\vspace{-0.8em}
\caption{Example where both models answer correctly on a prompt from the Countdown dataset.}
\label{fig:qualitative_correct_appendix}
\end{figure}

\begin{figure}[t]
\centering
\tiny
\setlength{\tabcolsep}{6pt}

\begin{tcolorbox}[
    colback=gray!3,
    colframe=gray!50,
    boxrule=0.5pt,
    arc=2pt,
    title={Qualitative comparison of failure cases (Qwen3-4B, GSM8K)}
]

\textbf{Prompt.}
Susan orders 3 magazines that send 12 issues a year.  She has 1 magazine that sends 6 issues a year.  Her last magazine sends her 4 times the amount of the 6 issue magazine.  How many magazines does she get every year?

\textbf{Correct answer:}
66
\vspace{0.6em}

\begin{tabular}{p{0.46\linewidth} p{0.46\linewidth}}
\textbf{Baseline (Qwen3-4B)} &
\textbf{TRSD (GSM8K)} \\[0.4em]

\begin{minipage}[t]{\linewidth}
\textbf{Think:}\\
Okay, let's see. Susan has three magazines. Wait, the problem says she orders 3 magazines that send 12 issues a year. Then there's another one that sends 6 issues a year. And her last magazine sends 4 times the amount of the 6-issue magazine. Hmm, need to figure out how many magazines she gets every year.

Wait, first, let me parse the question again. The first part says Susan orders 3 magazines that send 12 issues a year. So that's three magazines, each with 12 issues. Then she has 1 magazine that sends 6 issues a year. So that's another magazine. Then her last magazine sends her 4 times the amount of the 6-issue magazine. So the last magazine is 4 times 6, which is 24 issues. 

Wait, but the question is asking how many magazines does she get every year? Wait, maybe I need to check if the question is asking for the total number of magazines or the total number of issues? Wait, the wording says "how many magazines does she get every year?" So maybe it's the total number of magazines she has? Let me check again.

Original question: "How many magazines does she get every year?" So the answer is the total number of magazines she has. Let me check the problem again. Susan orders 3 magazines that send 12 issues a year. Then she has 1 magazine that sends 6 issues a year. Then her last magazine sends 4 times the amount of the 6-issue magazine. So, the total number of magazines would be 3 + 1 + 1 = 5? Because the first three are 12 issues each, then one is 6, and another one is 4 times 6, which is 24. So total magazines: 3 + 1 + 1 = 5. But wait, maybe the wording is tricky. Let me check again.

Wait, the problem says she orders 3 magazines that send 12 issues a year. Then she has 1 magazine that sends 6 issues. Then her last magazine sends 4 times the amount of the 6-issue magazine. So the last magazine is another one. So total number of magazines is 3 + 1 + 1 = 5. But maybe the problem is asking for the total number of issues? Wait, but the question says "how many magazines does she get every year?" So the answer is 5. But let me check again.

Alternatively, maybe the problem is phrased in a way that the three magazines are part of the total. Let me re-express:

Original problem:

Susan orders 3 magazines that send 12 issues a year. She has 1 magazine that sends 6 issues a year. Her last magazine sends her 4 times the amount of the 6 issue magazine. How many magazines does she get every year?

So, the first part: 3 magazines (each 12 issues). Then she has 1 magazine (6 issues). Then her last magazine is 4 times the 6-issue magazine. So that's 4*6=24 issues. So the last magazine is another one. So total number of magazines is 3 + 1 + 1 = 5. Therefore, the answer is 5. But let me check again if I'm misinterpreting the problem.

Alternatively, maybe the "last magazine" is part of the three she ordered? Wait, no. The problem says she orders 3 magazines that send 12 issues a year. Then she has 1 magazine that sends 6 issues. Then her last magazine sends 4 times the 6-issue. So that's an additional magazine. So total is 3 + 1 + 1 = 5. Therefore, the answer is 5. But maybe the problem is asking for the total number of issues? Wait, the question says "how many magazines does she get every year?" So it's the number of magazines, not issues. So answer is 5. But let me check again.

Wait, maybe the problem is in the wording. The first sentence says she orders 3 magazines that send 12 issues a year. Then she has 1 magazine that sends 6 issues. Then her last magazine sends 4 times the amount of the 6 issue magazine. So total magazines: 3 + 1 + 1 = 5. So the answer is 5. So I think that's correct. But maybe the problem is trying to trick me? Let me check again.

Alternatively, maybe the "last magazine" is part of the three she ordered? But the problem says she has 1 magazine that sends 6 issues. So that's separate. Then the last magazine is another one. So total of 3 + 1 + 1 = 5. Yes. Therefore, the answer is 5.

\vspace{0.4em}
\textbf{Answer:}
5
\end{minipage}
&
\begin{minipage}[t]{\linewidth}
\textbf{Think:}\\
Okay, let's break this down. Susan has three types of magazines. The first part says she orders 3 magazines that send 12 issues a year each. Then there's one magazine that sends 6 issues a year. The last magazine sends 4 times the amount of the 6-issue magazine. 

So, the first three magazines: 3 magazines * 12 issues = 36 issues. The next one is 6 issues. The last one is 4 times 6, which is 24 issues. Adding those together: 36 + 6 + 24 = 66. Wait, but the question asks how many magazines she gets every year. Wait, the total number of magazines is 3 + 1 + 1 = 5? Wait, no. Wait, the problem says "how many magazines does she get every year?" Wait, the issues per magazine, but the question is about the number of magazines. Wait, the problem says she orders 3 magazines that send 12 issues each, then 1 that sends 6, and one that sends 24 (4 times 6). So total number of magazines is 3 + 1 + 1 = 5. Wait, but the answer is 5? But the problem says "every year" so maybe the answer is 5. But let me check again.

The first part: 3 magazines that send 12 issues a year. Then one magazine that sends 6, and another that sends 4 times 6 (which is 24). So the total number of magazines is 3 + 1 + 1 = 5. So the answer is 5.

\vspace{0.4em}
\textbf{Answer:}
 5
\end{minipage}
\end{tabular}

\end{tcolorbox}

\vspace{-0.8em}
\caption{Example where both baseline and TRSD-trained models fail on the same prompt, illustrating a
shared failure mode.}
\label{fig:qualitative_wrong_1}
\end{figure}
\clearpage

\end{document}